\pdfoutput=1
\documentclass[10pt,twocolumn,letterpaper]{article}

\usepackage[pagenumbers]{cvpr}

\usepackage{cuted}
\usepackage{currfile}
\usepackage{caption}
\usepackage{placeins}
\usepackage{stfloats}

\renewcommand{\textfraction}{0.2}
\newcommand{\cvprfloatlimits}{%
  \setcounter{topnumber}{99}\setcounter{bottomnumber}{99}\setcounter{totalnumber}{99}%
  \setcounter{dbltopnumber}{2}\setcounter{dblbotnumber}{2}\renewcommand{\textfraction}{0.01}}

\definecolor{cvprblue}{rgb}{0.21,0.49,0.74}
\usepackage[pagebackref,breaklinks,colorlinks,allcolors=cvprblue]{hyperref}
\newcommand{\minisection}[1]{\noindent{\textbf{#1}.}}

\newcommand{\method}{HeadLight\xspace}
\newcommand{\zerowam}{Zero-WAM\xspace}
\newcommand{\robotwin}{RoboTwin\xspace}
\newcommand{\moheads}{Motion Heads\xspace}
\newcommand{\fastwam}{Fast-WAM\xspace}
\newcommand{\motus}{Motus\xspace}
\newcommand{\openwam}{OpenWAM-$\alpha$\xspace}
\newcommand{\cosmosedge}{Cosmos3-Edge\xspace}
\newcommand{\libero}{LIBERO\xspace}

\newcommand{\Kni}{24}
\newcommand{\Kmo}{48}

\newcommand{\mrh}[1]{\begin{tabular}[b]{@{}c@{}}#1\end{tabular}}
\newcommand{\ms}[2]{$#1{\pm}#2$}
\newcommand{\msb}[2]{$\mathbf{#1}{\pm}#2$}

\makeatletter
\patchcmd{\@maketitle}{\vspace*{12pt}}{\vspace*{1pt}}{}{}
\patchcmd{\endabstract}{\vspace*{12pt}}{\par}{}{}
\makeatother

\title{Targeted Adaptation of World-Action Models via Interaction and Motion Heads}

\author{
Yuze Zhu$^{1}$ \quad Jacky Li$^{2}$ \quad Yusen Luo$^{3}$ \quad Ken Jiankun Zheng$^{1}$\\
Chancharik Mitra$^{4}$ \quad Trevor Darrell$^{1}$ \quad Roei Herzig$^{1}$\\
{\normalsize\begin{tabular}[t]{@{}c@{}}$^{1}$University of California, Berkeley \quad $^{2}$University of Toronto\\
$^{3}$University of Southern California \quad $^{4}$Massachusetts Institute of Technology\end{tabular}}\\[4pt]
{\normalsize\url{https://headlight-wam.github.io/}}
}

\begin{document}
\maketitle
\begin{strip}
\centering
\includegraphics[width=\linewidth]{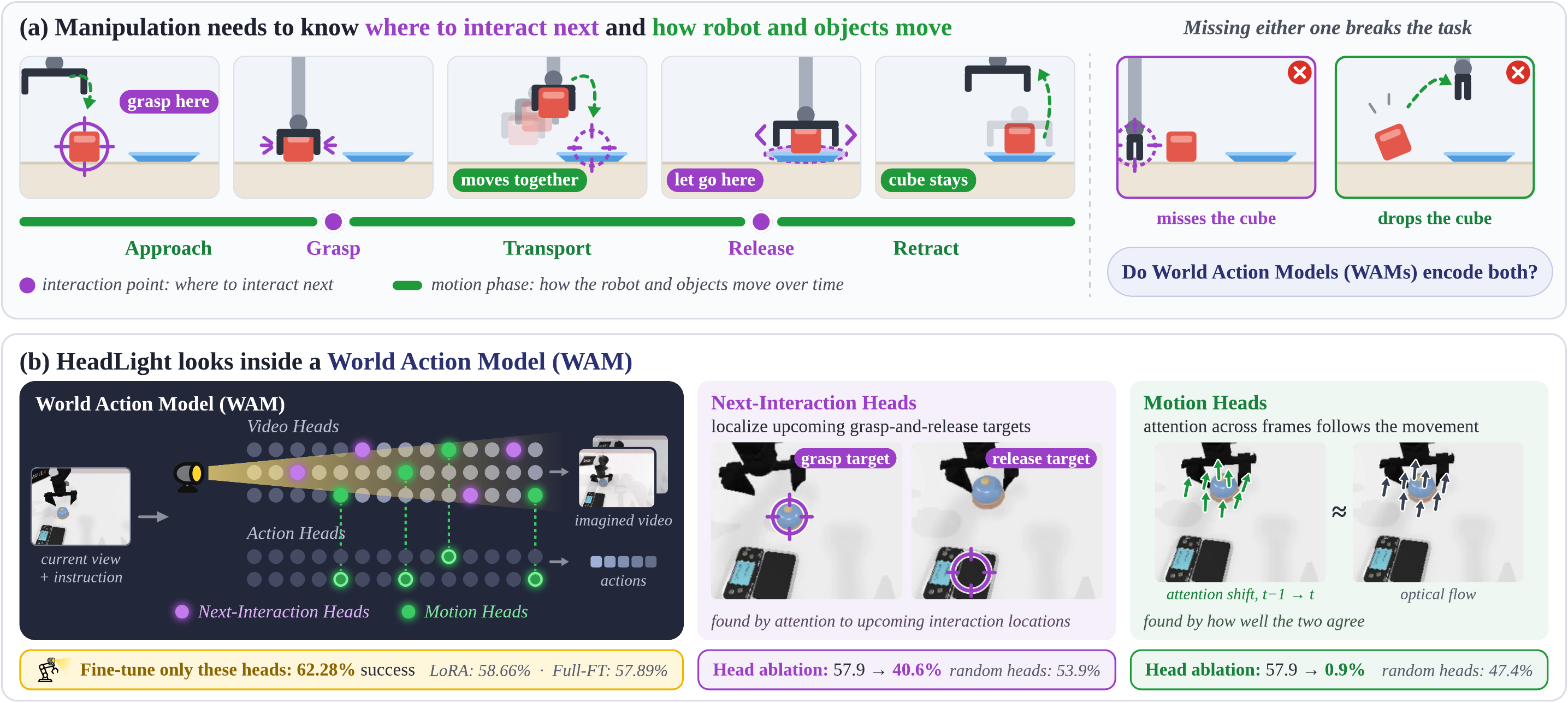}
\captionsetup{hypcap=false}
\captionof{figure}{\textbf{HeadLight: Targeted adaptation of WAMs via specialized attention heads.}
  (a)~Successful manipulation requires both localizing \emph{where to interact next} and modeling \emph{how the robot and objects move}; missing either breaks the task.
  (b)~Within pretrained WAMs, we identify \emph{Next-Interaction Heads} that localize upcoming grasp-and-release targets and \emph{Motion Heads} whose cross-frame attention aligns with optical flow.
  Ablating either head group substantially reduces manipulation success compared with equally sized random-head controls.
  \textbf{\method} fine-tunes only these heads, outperforming LoRA and full fine-tuning on unseen \robotwin tasks.}
\captionsetup{hypcap=true}
\label{fig:\currfilebase}%
\end{strip}

\begin{abstract}
World-Action Models (WAMs) are a promising new paradigm of robot manipulation policies that jointly generate future video and actions.
Successful manipulation requires both identifying where to interact and modeling how the robot and objects move.
Yet standard adaptation methods, such as full fine-tuning and LoRA, do not explicitly select parameters aligned with these manipulation-specific capabilities.
Motivated by this challenge, we discover two functionally specialized families of attention heads: Next-Interaction Heads, which localize upcoming grasp-and-release targets, and Motion Heads, which track movement across frames.
Furthermore, we introduce \method, a parameter-efficient refinement method that jointly adapts Next-Interaction Heads and Motion Heads.
On RoboTwin tasks unseen during fine-tuning, \method achieves 62.28\% success, compared with 57.89\% for full fine-tuning and 58.66\% for LoRA, while updating only 1.1\% of parameters.
Inference-time ablations further validate the functional importance of these heads: ablating Motion Heads reduces manipulation success from 57.89\% to 0.94\%, while ablating Next-Interaction Heads lowers it to 40.61\%, both substantially more than ablating equally many random heads.
Finally, we show that our approach is effective across multiple WAM architectures and real-robot manipulation settings.
\end{abstract}

\clearpage
\section{Introduction}

Robot manipulation has been explored through various policy architectures~\cite{chi2023diffusion,zitkovich2023rt2,black2025pi0}, with World-Action Models (WAMs) emerging as a promising approach that generates both future video and actions~\cite{ye2026dreamzero, zhou2026zerowam, kim2026cosmospolicy}. Research in motor control and task-and-motion planning describes manipulation in terms of discrete contact events and continuous motion~\cite{flanagan2006control, garrett2021tamp}. These aspects translate into two complementary capabilities that WAMs need: anticipating where the next grasp or release should occur, and representing the continuous motion of the robot and manipulated objects across frames.

Current WAMs implicitly rely on large-scale video pretraining~\cite{wan2025wan} to acquire these capabilities. Such pretraining allows them to generate visually plausible videos, yet \textit{visual} plausibility alone does not ensure that the \textit{manipulation} capabilities are actually learned~\cite{motamed2026physicsiq,qiu2026agra}. In practice, WAMs may fail to identify the correct interaction targets, and their generated actions may not accurately follow the motion in their predicted videos. These failures are especially pronounced when generalizing to unseen tasks. Although full fine-tuning and LoRA~\cite{hu2022lora} can adapt WAMs, neither method explicitly selects parameters based on their roles in interaction targeting and motion. As a result, optimizing the prediction loss may encourage the model to fit visual details, such as appearance and texture, without necessarily improving interaction targeting and motion modeling. This raises two natural questions: Can we identify the model components associated with interaction targeting and motion? If so, can we selectively refine them to strengthen these capabilities rather than merely fitting visual details?

Previous work has shown that robot policies can be improved by adapting a small set of internal components, either through steering sparse features at inference~\cite{zhou2026steeract} or by fine-tuning attention heads selected from demonstrations of a specific task~\cite{mitra2026robotsteering}. In both cases, the identified components are selected separately for each task. We instead ask whether some heads support more fundamental capabilities shared across manipulation tasks, namely interaction targeting and motion. To answer this, we investigate attention heads in pretrained WAMs and identify two families associated with these capabilities: \textbf{Next-Interaction Heads} and \textbf{Motion Heads}. We identify the former by analyzing attention to upcoming interaction locations, and the latter by measuring the agreement between cross-frame attention and optical flow. Ablating Motion Heads reduces manipulation success from 57.89\% to just 0.94\%, while ablating Next-Interaction Heads reduces it to 40.61\%, with both causing substantially larger declines than ablating equally many random heads. These experiments also reveal distinct failure modes: ablating Next-Interaction Heads often causes the robot to miss its intended targets, while ablating Motion Heads disrupts the motion in predicted videos. These results suggest that WAMs rely on a small number of specialized attention heads to localize interaction targets and to model motion.

Building on these findings, we introduce \textbf{\method}, a parameter-efficient refinement method that uses these two head families to guide complementary adaptation pathways. We selectively fine-tune Next-Interaction Heads to improve target localization and use Motion Heads to guide the adaptation of action prediction, helping generated actions follow the predicted motion. By combining these two pathways, our method aims to improve both interaction targeting and motion execution without updating the entire model.

We evaluate \method on RoboTwin~\cite{chen2025robotwin} tasks unseen during fine-tuning. On Zero-WAM, our method achieves \textbf{62.3\%} success, compared to 57.9\% for full fine-tuning and 58.7\% for LoRA, while updating only 1.1\% of parameters. Using either adaptation pathway alone does not achieve the same improvement, supporting the importance of their combination. We further demonstrate the effectiveness of our approach on LingBot-VA~\cite{li2026causalwam}, which uses a different WAM architecture. Finally, we validate in real-robot manipulation using a Franka robot.

To summarize, our main contributions are threefold: (i) We identify Next-Interaction Heads and Motion Heads in pretrained WAMs and validate their distinct roles in interaction targeting and motion through attention analysis and controlled ablations. (ii) We introduce \method, which combines two complementary adaptation pathways guided by these heads to improve interaction targeting and motion execution while updating only a small subset of model parameters. (iii) We demonstrate improvements over standard adaptation methods on unseen manipulation tasks, with effectiveness across WAM architectures and in real-robot manipulation. Together, these findings suggest that the functional structure of WAMs can guide which parameters to adapt, improving generalization to unseen tasks.

\section{Related Work}
\label{sec:related}
\begin{figure*}[t]
  \centering
  \includegraphics[width=\linewidth]{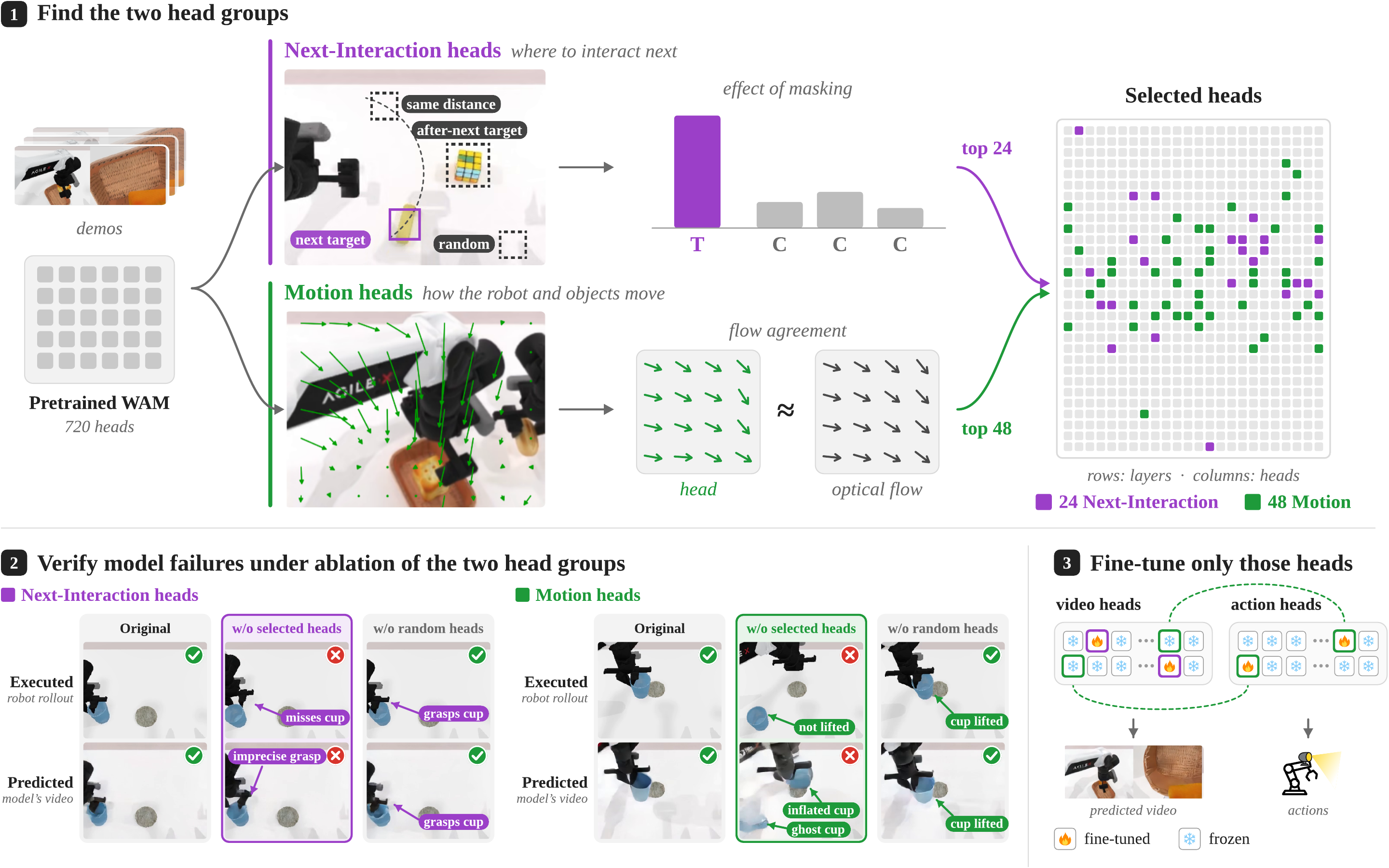}
  \caption{\textbf{Overview of \method.}
    (1) We identify Next-Interaction Heads by their selective reliance on upcoming interaction targets over control tokens, and Motion Heads by the alignment of cross-frame attention with optical flow.
    (2) Ablating these heads causes distinct failures: missed grasps and disrupted object motion, respectively, whereas ablating equally many random heads largely preserves successful execution.
    (3) \method selectively fine-tunes 24 Next-Interaction Heads in the video stream and 48 action heads paired with frozen video Motion Heads, while keeping all other heads frozen.}
  \label{fig:\currfilebase}
\end{figure*}

\paragraph{World-action models.}
World-action models jointly model future visual observations and robot actions for manipulation~\cite{guo2024pad,kim2026cosmospolicy,li2025uva,li2026causalwam,wu2024gr1,ye2026dreamzero,zhu2025uwm}.
Among these, Zero-WAM conditions its predictions on human demonstration videos to specify unseen tasks~\cite{zhou2026zerowam}.
While these models focus on learning to predict video and actions, we investigate the functional structure of their attention heads and how it can guide parameter selection for adaptation.

\paragraph{Functional specialization of attention heads.}
Prior studies have revealed functional specialization in language transformers, where attention heads capture linguistic relations~\cite{voita2019specialized,clark2019bert} and task representations during in-context learning~\cite{todd2024function}.
Similar properties have been explored in video diffusion models: DiTFlow extracts patch correspondences from cross-frame attention for motion transfer~\cite{pondaven2025ditflow}, while HALO identifies motion- and structure-sensitive heads, validates attention-derived displacements against optical flow, and leverages them for training-free motion transfer~\cite{jung2026halo}.
In embodied models, EWAM further reveals a layer-wise progression from semantic information to visual foresight and action~\cite{wang2026ewam}.
Building on these observations, we identify Next-Interaction Heads and Motion Heads in WAMs and examine their functional importance for closed-loop manipulation through controlled ablations.

\paragraph{Head selection, steering, and adaptation.}
Existing approaches adapt pretrained models through parameter updates or activation interventions.
LoRA~\cite{hu2022lora} constrains the rank of parameter updates without explicitly specifying which heads to adapt.
For robot policies, activation steering modifies behavior through interventions on feedforward activations~\cite{haon2025vlasteering}, while SteerAct~\cite{zhou2026steeract} selects a sparse feature from successful and failed rollouts and modulates its activation during execution.
Other approaches focus on attention heads: Sparse Attention Vectors~\cite{mitra2025savs} use few-shot examples to select attention-head activations for vision-language classification without fine-tuning, while Robotic Steering~\cite{mitra2026robotsteering} ranks heads using nearest-neighbor action regression on task demonstrations and selectively fine-tunes them with LoRA.
In contrast to task-specific head selection, HeadLight selects heads based on manipulation functions shared across tasks: selective reliance on upcoming interaction targets and alignment of cross-frame attention with optical flow.

\section{The \texorpdfstring{\method}{HeadLight} Method}
\label{sec:method}

We investigate whether functional specialization exists within pretrained WAMs and can guide parameter selection, aiming to strengthen interaction targeting and motion execution rather than fitting task-irrelevant visual details. First, given a frozen WAM, we identify two functionally specialized families of video attention heads, \emph{Next-Interaction Heads} and \emph{Motion Heads}, by scoring each head on how much it depends on the upcoming interaction target and how well its attention follows the motion of the scene (\cref{sec:discovery}). Second, we introduce \textbf{\method}, which uses these identified heads to establish two complementary, parameter-efficient adaptation targets: Next-Interaction Heads for localizing the upcoming interaction target, and Motion Heads for action execution (\cref{sec:refine}).

\subsection{Preliminaries: Joint Attention in WAMs}
\label{sec:prelim}

A WAM jointly generates future videos and actions via a transformer~\cite{vaswani2017attention} operating on both video and action tokens. At layer $\ell$ and head index $h$, we denote the attention head processing video tokens as a \emph{video head} $(\ell,h)_v$ and its counterpart for action tokens as an \emph{action head} $(\ell,h)_a$. The two modalities interact through \emph{joint attention}~\cite{liang2025mot}: at every layer, each action head $(\ell,h)_a$ attends to video tokens by querying the keys and values from its corresponding video head $(\ell,h)_v$. This one-to-one correspondence allows us to select target action heads through their paired video heads.

\subsection{Identifying Manipulation-Relevant Heads}
\label{sec:discovery}

Given a frozen WAM, we evaluate all video heads $(\ell,h)_v$ over manipulation episodes along two axes corresponding to discrete contact events and continuous trajectories: (i) selective reliance on the upcoming interaction target, and (ii) cross-frame alignment with scene motion.

\minisection{Next-Interaction Heads: Locating Where to Interact}
As a manipulation task progresses, the task-relevant spatial target shifts dynamically from the object to be grasped to its intended release location. A true \emph{Next-Interaction Head} should selectively track this \emph{upcoming} contact region rather than merely attending to visually salient objects or static background features.

We identify such heads via causal intervention: for each head, we mask out the tokens covering the upcoming interaction target and measure how much the head's output changes. Intuitively, a head that tracks the next interaction should be disrupted far more by this removal than by removing other plausible but task-irrelevant regions. We formalize this intuition below by comparing the target-removal effect against several control sets and retaining the heads with the largest margin.

Specifically, let $\mathcal{T}$ denote the video token that covers the upcoming interaction target in frame $f$. We ablate $\mathcal{T}$ from head $(\ell,h)_v$ by zeroing out its attention weights and re-normalizing the remaining distribution~\cite{geva2023dissecting}, then quantify the relative change in the head's output:
\begin{equation}
  D_{\ell,h}(\mathcal{T})
  =
  \frac{\left\| O_{\ell,h}^{v} - O_{\ell,h}^{v,\setminus\mathcal{T}} \right\|_2^2}{\left\| O_{\ell,h}^{v} \right\|_2^2},
  \label{eq:remove_change}
\end{equation}
where $O_{\ell,h}^{v}$ and $O_{\ell,h}^{v,\setminus\mathcal{T}}$ denote the head's outputs before and after removing $\mathcal{T}$, averaged across all video queries in frame $f$. A larger $D_{\ell,h}(\mathcal{T})$ indicates a stronger causal reliance on $\mathcal{T}$. However, a large $D_{\ell,h}(\mathcal{T})$ can arise for reasons other than next-interaction tracking: removing \emph{any} token perturbs a head's output, and the amount of perturbation depends on how much attention the token receives and how different its value is from the other tokens.

\begin{figure}[t]
  \centering
  \includegraphics[width=\columnwidth]{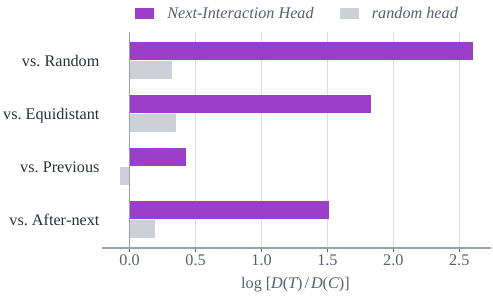}
  \caption{\textbf{Selectivity of Next-Interaction Heads.}
    Log-ratio margin (\cref{eq:score_ni}) between masking the upcoming target $\mathcal{T}$ and masking each of four control tokens $\mathcal{C}$: Random, Equidistant (as far from the gripper as $\mathcal{T}$), Previous (the previous target), and After-next (the target after $\mathcal{T}$).
    A Next-Interaction Head maintains positive margins across all controls, whereas a same-layer random head stays near zero.}
  \label{fig:\currfilebase}
\end{figure}

\begin{figure}[t]
  \centering
  \includegraphics[width=\columnwidth]{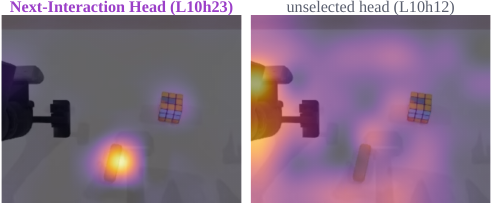}
  \caption{\textbf{Target-selective attention of a Next-Interaction Head.}
    Spatial attention maps from the same layer as the left gripper approaches the target.
    The Next-Interaction Head focuses sharply on the upcoming grasp target while ignoring the distractor object, whereas an unselected head attends diffusely across the scene.}
  \label{fig:\currfilebase}
\end{figure}

To rule out such alternative explanations, we apply the same removal to four control tokens $\mathcal{C}^{c}$ ($c \in \{1,\dots,4\}$) from the same frame and compute $D_{\ell,h}(\mathcal{C}^{c})$ with \cref{eq:remove_change}. Each control isolates an alternative explanation: if removing it disrupts the head as much as removing $\mathcal{T}$, that explanation cannot be excluded. The controls are (1) a \emph{random} token, ruling out that the head is sensitive to arbitrary removals; (2) an \emph{equidistant} token, as far from the gripper as $\mathcal{T}$, ruling out that the head merely attends near the gripper; (3) the \emph{previous} interaction target, ruling out that it tracks where the robot has already been; and (4) the \emph{after-next} interaction target, ruling out that it attends to all task-relevant targets rather than only the next.

Each control yields a log-ratio margin, $\log\big(\mathbb{E}[D_{\ell,h}(\mathcal{T})]/\mathbb{E}[D_{\ell,h}(\mathcal{C}^{c})]\big)$, which is positive when removing $\mathcal{T}$ disrupts the head more than removing that control. A head supports next-interaction tracking only if its margin is positive against all four controls, so we define its selectivity score as the smallest of the four margins:
\begin{equation}
  S_{\mathrm{N}}(\ell,h)
  =
  \min_{c}\,
  \log
  \frac{
    \mathbb{E}_{f}\big[D_{\ell,h}(\mathcal{T})\big]
  }{
    \mathbb{E}_{f}\big[D_{\ell,h}(\mathcal{C}^{c})\big]
  },
  \label{eq:score_ni}
\end{equation}
where $\mathbb{E}_f$ averages over frames across manipulation episodes. $S_{\mathrm{N}}(\ell,h)$ is therefore high only if the head strictly relies on the upcoming target more than on all four controls. Finally, we rank all video heads by $S_{\mathrm{N}}(\ell,h)$ and select the top $K_{\mathrm{N}}=\Kni$ as the \emph{Next-Interaction Heads} $\mathcal{H}_{\mathrm{N}}^{v}$, whose selectivity and attention are illustrated in \cref{fig:ni_selectivity,fig:ni_attention}.

\begin{figure}[t]
  \centering
  \includegraphics[width=\columnwidth]{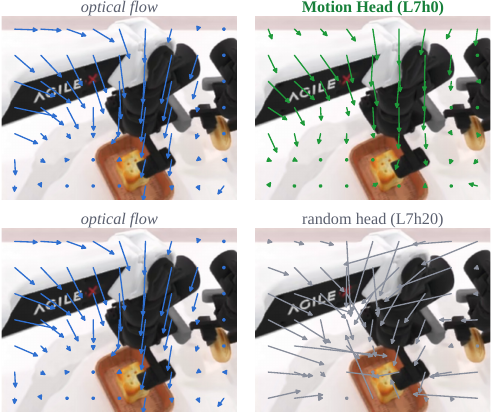}
  \caption{\textbf{Cross-frame motion alignment of Motion Heads.}
    Reference optical flow $\mathbf{u}^{f}$ (left) versus attention-induced motion fields $\widehat{\mathbf{u}}_{\ell,h}^{f}$ (\cref{eq:attn_flow}, right).
    A Motion Head closely tracks the moving robot arm and gripper, whereas a same-layer random head yields disorganized offsets.}
  \label{fig:\currfilebase}
\end{figure}

\minisection{Motion Heads: Capturing How the Scene Moves}
While Next-Interaction Heads identify where the robot should interact, successful manipulation also requires understanding how the robot and objects move over time. We therefore seek attention heads whose internal attention patterns align with the physical motion observed in the scene. Motivated by previous observations that cross-frame attention can encode motion correspondences~\cite{pondaven2025ditflow,nam2025difftrack,jung2026halo}, we identify such heads by measuring how closely their attention-induced motion fields align with reference optical flow. This distinguishes heads whose cross-frame attention moves with the scene from heads whose attention stays at fixed image locations or is unrelated to the motion.

To measure this alignment, we first derive an attention-induced motion field for each video head $(\ell,h)$. For a spatial token $p$ at coordinate $\mathbf{x}_p$ in frame $f$, we consider its spatial offsets relative to tokens $p'$ in the preceding frame $f-1$, and aggregate them according to the head's cross-frame attention weights:
\begin{equation}
    \widehat{\mathbf{u}}_{\ell,h}^{f}(p)
    =
    \sum_{p'}
    \bar{A}_{\ell,h}^{f\rightarrow f-1}(p,p')
    \left(\mathbf{x}_p-\mathbf{x}_{p'}\right),
    \label{eq:attn_flow}
\end{equation}
where $\bar{A}_{\ell,h}^{f\rightarrow f-1}(p,p')$ denotes the attention weight of token $p$ attending to token $p'$, normalized over all spatial tokens $p'$ in frame $f-1$. As visualized in \cref{fig:motion_flow}, this produces a dense, patch-wise displacement field that characterizes the motion correspondences encoded by each head.

We then quantify the alignment between this attention-induced field and reference optical flow $\mathbf{u}^{f}$~\cite{kroeger2016dis}, using the expected cosine similarity:
\begin{equation}
    S_{\mathrm{M}}(\ell,h)
    =
    \mathbb{E}_{f,p}
    \Big[
        \cos\big(
            \widehat{\mathbf{u}}_{\ell,h}^{f}(p),
            \mathbf{u}^{f}(p)
        \big)
    \Big].
    \label{eq:score_motion}
\end{equation}
We rank all video heads by $S_{\mathrm{M}}(\ell,h)$ and select the top $K_{\mathrm{M}}=\Kmo$ as the \emph{Motion Heads} $\mathcal{H}_{\mathrm{M}}^{v}$, with both selected head sets shown in \cref{fig:method_overview}. We use $K_{\mathrm{N}}=\Kni$ and $K_{\mathrm{M}}=\Kmo$ throughout and analyze these budgets in \cref{sec:ablations}.

\paragraph{Functional validation of the identified heads.}
While $S_{\mathrm{N}}$ and $S_{\mathrm{M}}$ measure target selectivity
and motion alignment, they do not establish whether the identified
heads are necessary for closed-loop manipulation.
We therefore conduct inference-time head ablations against
equally sized random-head controls in \cref{sec:exp_causal},
where \cref{fig:head_ablation} confirms their distinct functional importance for interaction
targeting and motion execution.
\subsection{Selective Refinement of Interaction and Motion}
\label{sec:refine}

Once $\mathcal{H}_{\mathrm{N}}^{v}$ and $\mathcal{H}_{\mathrm{M}}^{v}$ are identified, \method reuses them across downstream tasks for selective fine-tuning.
On new tasks, target errors originate from predicted videos missing the right object or placement location, whereas motion errors typically arise when generated actions fail to follow a plausible video trajectory.
\method accordingly pairs two complementary pathways: refining $\mathcal{H}_{\mathrm{N}}^{v}$ in the video stream for target localization, while keeping $\mathcal{H}_{\mathrm{M}}^{v}$ frozen and refining its paired action heads for motion execution.

\paragraph{Video pathway: refining Next-Interaction Heads.}
To sharpen where upcoming grasps and releases appear in predicted videos, we selectively fine-tune the $K_{\mathrm{N}}$ Next-Interaction Heads $\mathcal{H}_{\mathrm{N}}^{v}$ while freezing all other video heads.
Restricting video updates to $\mathcal{H}_{\mathrm{N}}^{v}$ focuses adaptation on localizing task-relevant contact targets, rather than fitting task-specific visual details such as background appearance and texture.

\begin{table*}[t]
  \centering
  \setlength{\tabcolsep}{3pt}
  \renewcommand{\arraystretch}{1.1}
  \caption{\textbf{Generalization to unseen \robotwin~\cite{chen2025robotwin} tasks across WAM architectures.}
    Success rate (\%, mean${\pm}$std over 3 seeds, 100 trials each) after adapting on 43 seen tasks; unadapted backbone results are from~\cite{zhou2026zerowam}.
    \method outperforms Full FT and LoRA on LingBot-VA and \zerowam.
    Best per backbone in bold.}
  \label{tab:\currfilebase}
  \begin{tabular*}{\linewidth}{@{\extracolsep{\fill}}lccccccc@{}}
    \toprule
    Method
      & \mrh{Place object\\on scale} & \mrh{Stamp\\seal} & \mrh{Move stapler\\to pad}
      & \mrh{Place bread\\in basket} & \mrh{Place\\empty cup} & \mrh{Stack\\blocks three} & Average \\
    \midrule
    LingBot-VA~\cite{li2026causalwam}
      & \ms{6.17}{4.87} & \ms{3.67}{2.49} & \ms{23.33}{8.22}
      & \ms{17.33}{6.18} & \ms{42.33}{7.85} & \ms{0.00}{0.00} & \ms{15.47}{1.30} \\

    \quad + Full FT
      & \ms{22.33}{3.77} & \ms{21.67}{2.87} & \ms{61.00}{4.24}
      & \ms{34.67}{4.92} & \ms{56.00}{4.32} & \ms{0.00}{0.00} & \ms{32.61}{3.04} \\
    \quad + LoRA
      & \ms{20.67}{3.09} & \ms{21.33}{2.49} & \ms{64.00}{2.94}
      & \ms{37.33}{1.25} & \ms{61.33}{4.50} & \ms{0.00}{0.00} & \ms{34.11}{1.31} \\
    \quad + \method
      & \msb{23.00}{7.07} & \msb{25.67}{2.36} & \msb{65.67}{4.50}
      & \msb{49.33}{2.62} & \msb{65.33}{0.47} & \ms{0.00}{0.00} & \msb{38.17}{1.11} \\
    \midrule
    \zerowam~\cite{zhou2026zerowam}
      & \ms{24.67}{2.05} & \ms{47.00}{4.55} & \ms{69.14}{2.93}
      & \ms{35.00}{3.74} & \ms{84.87}{0.18} & \ms{9.00}{2.16} & \ms{44.95}{0.79} \\
    \quad + Full FT
      & \ms{43.33}{1.25} & \msb{73.33}{6.65} & \ms{85.33}{0.47}
      & \ms{42.67}{2.87} & \ms{91.33}{2.05} & \msb{11.33}{3.09} & \ms{57.89}{0.77} \\

    \quad + LoRA
      & \ms{47.33}{5.79} & \ms{70.33}{3.30} & \ms{83.00}{2.94}
      & \ms{48.33}{2.62} & \ms{96.00}{0.82} & \ms{7.00}{2.16} & \ms{58.66}{1.87} \\
    \quad + \method
      & \msb{54.00}{5.72} & \msb{73.33}{3.40} & \msb{89.00}{1.41}
      & \msb{52.67}{4.03} & \msb{97.00}{1.63} & \ms{7.67}{1.70} & \msb{62.28}{2.30} \\
    \bottomrule
  \end{tabular*}
\end{table*}

\paragraph{Action pathway: refining Motion Heads.}
Pretrained Motion Heads $\mathcal{H}_{\mathrm{M}}^{v}$ already capture general scene motion reliably, so we keep $\mathcal{H}_{\mathrm{M}}^{v}$ frozen to preserve these priors and instead adapt how the action stream reads them.
Recall from \cref{sec:prelim} that each action head $(\ell,h)_a$ attends to video tokens by querying the keys and values of its paired video head $(\ell,h)_v$.
We therefore select the action heads at the same layer and head indices as $\mathcal{H}_{\mathrm{M}}^{v}$:
\begin{equation}
  \mathcal{H}_{\mathrm{M}}^{a}
  =
  \big\{
    (\ell,h)_a
    \;\big|\;
    (\ell,h)_v \in \mathcal{H}_{\mathrm{M}}^{v}
  \big\},
  \label{eq:action_heads}
\end{equation}
which, for brevity, we also refer to as \moheads, and fine-tune only these $K_{\mathrm{M}}$ heads in the action stream.
Because $\mathcal{H}_{\mathrm{M}}^{a}$ directly reads the motion features of $\mathcal{H}_{\mathrm{M}}^{v}$, adapting it helps generated actions follow predicted video trajectories without optical-flow supervision during fine-tuning.

\paragraph{Applicability across WAMs.}
\method generalizes to joint-attention WAMs with either separate modality projections or a shared backbone, including human-video-conditioned \zerowam~\cite{zhou2026zerowam} and language-conditioned LingBot-VA~\cite{li2026causalwam}.
It updates only a small fraction of parameters without incurring the additional inference overhead associated with activation steering methods~\cite{haon2025vlasteering,zhou2026steeract}.

\section{Experiments}

In this section, we evaluate whether the identified Next-Interaction Heads and Motion Heads genuinely capture interaction targeting and motion execution, and whether parameter-efficient adaptation through these functionally specialized attention heads improves these capabilities and generalization to unseen manipulation tasks. Our experiments address three key questions:  (i) Does selectively refining Next-Interaction Heads and Motion Heads outperform standard adaptation methods on unseen tasks across distinct WAM architectures? (ii) Are the identified heads causally necessary for their respective functions, with complementary benefits from their joint refinement? (iii) Do the performance improvements observed on unseen simulation tasks transfer to real-robot manipulation, where the model must generalize to unseen object--goal combinations from limited demonstrations?

\subsection{Experimental Setup}
\label{sec:exp_setup}

\noindent\textbf{Models and benchmarks.} To test whether HeadLight generalizes across WAM designs, we evaluate two distinct architectures: \textbf{Zero-WAM}~\cite{zhou2026zerowam}, which uses separate video and action heads conditioned on in-context human demonstration videos, and \textbf{LingBot-VA}~\cite{li2026causalwam}, whose released model processes future video and action tokens within a shared transformer conditioned on language instructions. For simulation, we follow the cross-task protocol of Zero-WAM~\cite{zhou2026zerowam} on RoboTwin~\cite{chen2025robotwin}. We fine-tune policies on 43 seen bimanual tasks and evaluate generalization on 6 unseen tasks, with 100 episodes × 3 seeds per task. We note that \emph{Open Microwave} has a simulator issue and is therefore excluded from our evaluation (see \cref{sec:supp_sim}). We highlight that unseen tasks provide a direct test of manipulation-specific adaptation because they require transferring interaction targeting and motion execution to novel goals rather than memorizing training scenes.

\noindent\textbf{Baselines and implementation.}
We compare HeadLight with the unadapted backbone, Full Fine-Tuning (Full FT), and LoRA~\cite{hu2022lora}.
All adaptation methods start from the same post-trained checkpoint and are trained on identical seen-task data for 2{,}000 steps on Zero-WAM and 10{,}000 steps on LingBot-VA.
For LoRA, we choose rank $r{=}77$ to match HeadLight's trainable parameter budget on Zero-WAM (113.5M vs.\ 113.3M), applying it to the self-attention projections of all video and action heads in both WAMs. Full FT updates all model parameters.
Full implementation, evaluation, and head-discovery details are provided in \cref{sec:supp_impl,sec:supp_discovery}.

\subsection{Generalization to Unseen Tasks Across WAMs}
\label{sec:main_results}

As shown in Tab.~\ref{tab:main_results}, HeadLight achieves the highest overall success rate of \textbf{62.28\%} on Zero-WAM, outperforming Full FT by 4.39 points and LoRA by 3.62 points. Unlike LoRA, which distributes its update capacity across attention heads without explicitly considering their functions, HeadLight adapts only the Next-Interaction and Motion Heads identified by our functional analysis. The higher success on unseen tasks suggests that selecting heads by function is an effective way to adapt WAMs.

We further evaluate HeadLight on LingBot-VA to assess its generalizability to a unified, language-conditioned WAM architecture. HeadLight consistently achieves the strongest performance, reaching \textbf{38.17\%} average success and outperforming Full FT by 5.56 points and LoRA by 4.06 points. These results suggest that functionally guided head adaptation generalizes across different WAM architectures.

\subsection{Ablation Studies}
\label{sec:ablations}
\label{sec:exp_causal}

\begin{figure}[t]
  \centering
  \includegraphics[width=\columnwidth]{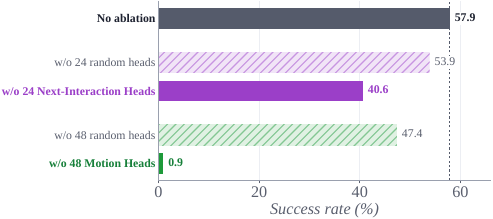}
  \caption{\textbf{Inference-time causal head ablation} on unseen \robotwin tasks (\zerowam after Full FT).
Lower success rates indicate stronger functional importance: ablating Next-Interaction Heads or Motion Heads substantially reduces manipulation success compared with equally sized random-head controls, confirming their critical roles in task execution.}
  \label{fig:\currfilebase}
\end{figure}

\noindent\textbf{Functional roles of the identified heads.}
At inference time, we zero out each head group~\cite{michel2019sixteen} on Zero-WAM to test their functional importance for successful manipulation.
As shown in \cref{fig:head_ablation}, ablating Next-Interaction Heads reduces success to 40.61\%, while ablating Motion Heads lowers it to 0.94\%, both substantially larger drops than ablating equally many random heads.
The two sets are disjoint among all 720 video heads, and ablating them leads to failure modes in \cref{fig:method_overview}: mislocalized grasps missing the target without Next-Interaction Heads, versus ghosted object motion and stalled execution without Motion Heads.

\begin{table}[t]
  \centering
  \caption{\textbf{Complementarity of the two adaptation pathways} on unseen \robotwin tasks (\zerowam).
We compare adapting Next-Interaction Heads, Motion Heads, and their combination against equally sized random-head controls.
Success rates are reported as mean${\pm}$std over 3 seeds, with 100 trials per seed.}
  \label{tab:\currfilebase}
  \small
  \setlength{\tabcolsep}{4pt}
  \begin{tabular}{@{}lrc@{}}
    \toprule
    Adapted heads & \#Params & SR (\%) $\uparrow$ \\
    \midrule
    None (\zerowam~\cite{zhou2026zerowam}) & -- & \ms{44.95}{0.79} \\
    All (Full FT) & 10.0B & \ms{57.89}{0.77} \\
    \midrule
    Next-Interaction Heads only & 37.8M & \ms{56.17}{0.14} \\
    \quad random heads control & 37.8M & \ms{52.22}{0.44} \\
    Motion Heads only & 75.5M & \ms{56.11}{1.36} \\
    \quad random heads control & 75.5M & \ms{54.33}{1.65} \\
    \midrule
    Both (random heads control) & 113.3M & \ms{57.33}{1.89} \\
    Both (\method) & 113.3M & \msb{62.28}{2.30} \\
    \bottomrule
  \end{tabular}
\end{table}

\noindent\textbf{Complementarity of the two adaptation pathways.}
To examine whether the two adaptation pathways provide complementary benefits, we compare each pathway individually with their combination in \cref{tab:complementarity}.
Adapting Next-Interaction Heads or Motion Heads alone outperforms adapting equally many random heads, with further gains achieved by combining both pathways in HeadLight, reaching $62.28\%$ success, surpassing Full FT ($57.89\%$) and outperforming either individual pathway by over six points.
These results demonstrate the complementary benefits of the two adaptation pathways.

\begin{figure}[htb]
  \centering
  \includegraphics[width=\columnwidth]{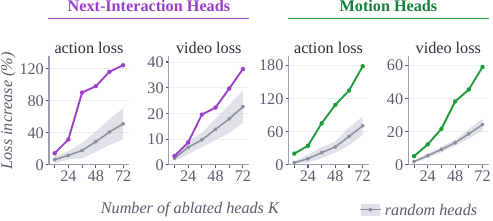}
  \caption{\textbf{Head-budget analysis via loss sensitivity.}
    Action- and video-loss increases from ablating top-$K$ heads versus random-head controls (gray, mean${\pm}$std) in \zerowam.
    Up to $K_{\mathrm{N}}{=}\Kni$, ablating Next-Interaction Heads increases action loss while leaving video loss near random-head controls; ablating Motion Heads substantially increases both losses across $K$.}
  \label{fig:\currfilebase}
\end{figure}

\noindent\textbf{Choice of head budgets.}
To determine how many heads are needed for each adaptation pathway, we examine how head budgets affect prediction losses and unseen-task success.
Specifically, we progressively ablate the top-$K$ heads and measure the resulting increases in action and video prediction losses against equally sized random-head controls, as shown in Fig.~\ref{fig:head_budget}.
For Next-Interaction Heads, the top 24 strongly affect action prediction while leaving video prediction largely intact; including more heads increasingly perturbs video prediction without improving unseen-task success (see \cref{sec:supp_budget}).
For Motion Heads, adapting their paired action heads improves success up to $K_{\mathrm{M}}{=}48$, while ablating these 48 video heads reduces success to near zero in Fig.~\ref{fig:head_ablation}, indicating that these heads already encode scene motion.
These results motivate selecting $K_{\mathrm{N}}{=}24$ for interaction targeting and $K_{\mathrm{M}}{=}48$ for motion-guided action adaptation, balancing functional specificity and adaptation effectiveness.

\begin{figure*}[t]
  \centering
  \includegraphics[width=\linewidth]{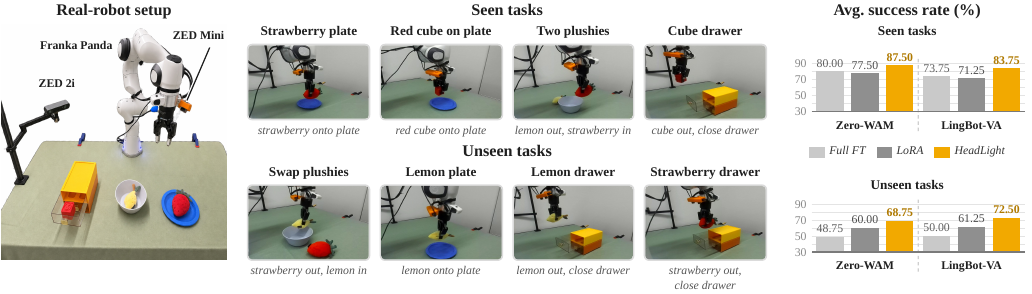}
  \caption{\textbf{Real-robot workspace, tasks, and results.}
  \emph{Left:} Franka Panda workspace with third-person ZED~2i and wrist-mounted ZED~Mini cameras.
  \emph{Middle:} Four seen adaptation tasks (top) and four unseen tasks that recombine objects and manipulation goals (bottom), each evaluated over 20 rollouts.
  \emph{Right:} Average success rate across four tasks per split.
  \method outperforms Full FT and LoRA on both WAMs, with larger gains on unseen tasks.}
  \label{fig:\currfilebase}
\end{figure*}

\subsection{Real-World Manipulation}
\label{sec:real_world}

To evaluate HeadLight beyond simulation, we conduct real-world experiments on the 7-DoF Franka Panda arm shown in Fig.~\ref{fig:real_setup}. We consider four seen tasks, each with 100 teleoperated demonstrations for adaptation, and four unseen tasks that recombine objects and manipulation goals in novel configurations. Each task is evaluated over 20 physical rollouts on both Zero-WAM and LingBot-VA, with per-task results reported in \cref{sec:supp_real_results}.

HeadLight consistently outperforms both Full FT and LoRA across all settings on both WAMs. Beyond achieving the highest success rates on seen tasks, its gains are especially pronounced on unseen tasks, reaching 68.75\% on Zero-WAM (+8.75 points over LoRA) and 72.50\% on LingBot-VA (+11.25 points over LoRA). These results demonstrate that HeadLight's benefits extend beyond the training tasks and suggest that functionally guided adaptation improves the transfer of interaction targeting and motion execution to novel manipulation goals.

\section{Conclusion}
We show that pretrained World-Action Models (WAMs) contain two disjoint families of functionally specialized attention heads: Next-Interaction Heads, which localize upcoming interaction targets, and Motion Heads, which capture scene motion across frames.
Through controlled inference-time ablations, we validate their distinct functional roles in manipulation, showing substantially larger performance drops than equally sized random-head controls.
Guided by this functional structure, we introduce HeadLight, a parameter-efficient refinement method with two complementary adaptation pathways.
It fine-tunes Next-Interaction Heads for interaction targeting and action heads paired with frozen video Motion Heads for motion execution.
Experiments across two distinct WAM architectures, Zero-WAM and LingBot-VA, demonstrate consistent gains over full fine-tuning and LoRA on unseen RoboTwin tasks and on real-robot manipulation with a Franka Panda arm.
Together, these findings suggest that functionally guided parameter selection can improve generalization by prioritizing manipulation-relevant capabilities over task-specific visual details.

\section{Limitations and Future Work}
\label{sec:limitations}
Our head discovery has two requirements. First, although the identified Next-Interaction Heads can be reused across unseen tasks, their initial discovery requires datasets annotated with upcoming interaction targets. Second, identifying Motion Heads relies on attention from the current frame to the preceding frame to construct attention-induced motion fields. Extending motion-based head discovery to architectures with different temporal attention mechanisms remains an interesting direction.

Beyond discovery, HeadLight uses these functional heads to guide selective fine-tuning. Their specialized roles open up further opportunities to improve interaction targeting and motion execution. For instance, auxiliary localization objectives could sharpen interaction targeting, while task-specific cues extracted from human demonstrations could provide more direct guidance to these heads. Alternatively, directly steering their activations could enable more flexible control over manipulation behavior without updating model parameters.

\section*{LLM Usage Statement}
Claude Code was used to assist with code development and programmatic figure creation. ChatGPT and Claude assisted with manuscript writing and refinement. GPT Image 2.6 was used solely for background removal in the real-robot setup image. All AI-assisted content was reviewed and verified by the authors.

{
    \small
    \bibliographystyle{ieeenat_fullname}
    \bibliography{main}
}

\cvprfloatlimits
\clearpage
\maketitlesupplementary
\appendix
\raggedbottom
\makeatletter
\setlength{\@fptop}{0pt}\setlength{\@fpsep}{\floatsep}\setlength{\@fpbot}{0pt plus 1fil}
\setlength{\@dblfptop}{0pt}\setlength{\@dblfpsep}{\dblfloatsep}\setlength{\@dblfpbot}{0pt plus 1fil}
\renewcommand{\floatpagefraction}{0.8}
\renewcommand{\dblfloatpagefraction}{0.8}
\g@addto@macro\@doclearpage{\if@twocolumn\if@firstcolumn\global\@colroom\@colht\global\vsize\@colht\fi\fi}
\makeatother

\noindent This supplementary material provides additional details and analyses supporting the main paper.
\cref{sec:supp_impl} describes model architectures, training, head-level fine-tuning, and the evaluation protocol.
\cref{sec:supp_heads} details the discovery procedure, reports causal ablations on \zerowam and four additional WAMs on \robotwin and \libero, and validates head consistency.
\cref{sec:supp_results} reports per-task breakdowns, head-budget analyses, and qualitative comparisons.
\cref{sec:supp_real} details real-robot hardware, tasks, and per-task results.
\cref{sec:supp_sim} documents a \robotwin simulator artifact that leads us to exclude \emph{Open Microwave} from evaluation.

\section{Implementation and Experimental Details}
\label{sec:supp_impl}

\subsection{Models and Training}
\label{sec:supp_train}

We adapt \method on \zerowam and LingBot-VA, which couple video and actions in different ways and both use 30 transformer layers with 24 attention heads of dimension 128 per layer.
\zerowam employs separate video and action streams coupled through joint attention and conditions its generation on an in-context human demonstration video.
In contrast, LingBot-VA processes future video and action tokens within a unified transformer with shared attention weights and conditions its generation on a language instruction.
Beyond these adaptation backbones, \cref{sec:supp_crosswam} applies head discovery and causal ablation to four more publicly released WAMs with different architectures, on \robotwin and on \libero, confirming that the analysis pipeline transfers across architectures.

All adaptation methods are trained on clean demonstrations from the 43 seen tasks of \robotwin.
\zerowam pairs each robot episode with a human video from the \zerowam dataset~\cite{zhou2026zerowam} and jointly optimizes video prediction, action prediction, and multi-chunk prediction.
LingBot-VA optimizes its standard video and action prediction objectives.
\method routes these objectives separately to its two adaptation pathways (\cref{sec:supp_headft}).

\paragraph{Starting checkpoints and optimization.}
For \zerowam, all methods start from the released RoboTwin post-trained checkpoint and train for 2{,}000 steps.
Because the publicly released LingBot-VA RoboTwin checkpoint is trained on all 50 tasks including our held-out evaluation tasks, we first adapt the released pretrained checkpoint to the 43 seen tasks to obtain a clean starting model.
All methods then train from this model for 10{,}000 steps.
All fine-tuning runs use the AdamW optimizer with $\beta{=}(0.9,0.95)$, zero weight decay, gradient clipping at $1.0$, and a 200-step linear warm-up.
On \zerowam, the learning rate then follows a cosine decay to $0.1{\times}$ the maximum; on LingBot-VA, it is held at the maximum after the warm-up and cosine-decayed to $0.1{\times}$ over the final $1{,}000$ steps.
Both LoRA and \method use a maximum learning rate of $10^{-4}$ with an effective batch size of 8 on \zerowam and LingBot-VA.
Following their respective standard recipes, Full FT uses a maximum learning rate of $2{\times}10^{-5}$ with batch size 8 on \zerowam and $10^{-5}$ with batch size 16 on LingBot-VA.
All experiments are conducted on 8 NVIDIA A100 or H100 GPUs.%

\subsection{Head-Level Selective Fine-Tuning}
\label{sec:supp_headft}

For each selected attention head $(\ell,h)$, \method introduces a zero-initialized additive update to the 128 output rows of the query, key, and value projections, including their weights and biases, together with the 128 input columns of the output projection.
Each selected head accounts for $1.57$M trainable parameters.
Adapting the $\Kni$ Next-Interaction Heads and $\Kmo$ Motion Heads yields $113.3$M trainable parameters in total, representing $1.1\%$ of \zerowam's $10.0$B parameters, while all remaining attention heads, feed-forward layers, normalizations, and embeddings stay frozen, substantially reducing training memory usage relative to Full FT.

Within each optimization step, the two adaptation pathways are updated by distinct loss signals from a single forward pass.
Specifically, gradients from the video prediction objectives update only the Next-Interaction Heads, whereas gradients from the action prediction objective update only the action heads paired with the frozen Motion Heads.
Because every head-level update is additive over existing linear projections, all updates can be merged directly into the base weights after training, leaving the model architecture and inference cost completely unchanged.

\paragraph{Adaptation on shared-weight WAMs.}
In the unified LingBot-VA architecture, video and action tokens share a single set of self-attention weights rather than separate modality streams.
To preserve the same dual-pathway routing, \method applies Next-Interaction Head updates exclusively to video tokens and Motion Head updates exclusively to action tokens, keeping separate updates per token type whenever a head belongs to both sets.
Since LingBot-VA executes separate video and action forward passes at inference time, each update merges cleanly into the corresponding pass with zero inference overhead.

\paragraph{Baselines.}
Full FT updates the entire WAM backbone under the standard training objective.
For LoRA, we apply rank $r{=}77$ with scaling $\alpha{=}r$ to the query, key, value, and output projections of self-attention across all video and action heads in \zerowam and LingBot-VA, trained with the combined loss.
This rank is chosen so that LoRA matches the trainable parameter budget of \method in \zerowam, with $113.5$M trainable parameters for LoRA versus $113.3$M for \method.

\subsection{Simulation Evaluation Protocol}
\label{sec:supp_eval}

We follow the cross-task evaluation protocol of \zerowam~\cite{zhou2026zerowam} on \robotwin~2.0~\cite{chen2025robotwin} using clean scenes and rasterized rendering.
Following \zerowam, each unseen task is evaluated over three seeds with 100 expert-solvable scenes per seed, and all methods run on identical scenes.
Each cell in \cref{tab:main_results} reports the mean and standard deviation across seeds; overall averages are computed per-seed before aggregating.

During closed-loop rollouts, the policy observes three camera views, predicts the next chunk of future video frames and end-effector actions, executes them, and re-plans until success or the episode limit.
\zerowam uses one fixed human demonstration video per task with 50 denoising steps and classifier-free guidance scale 5; LingBot-VA uses the task language instruction with its released inference settings.
The unadapted baselines in \cref{tab:main_results} follow the published results of~\cite{zhou2026zerowam}.
We exclude \emph{Open Microwave} due to a simulator success-check artifact.
The additional WAMs of \cref{sec:supp_crosswam} follow each model's official evaluation protocol.

\FloatBarrier

\section{Functional Head Discovery and Analysis}
\label{sec:supp_heads}

\subsection{Discovery Protocol}
\label{sec:supp_discovery}

\begin{figure}[t]
  \centering
  \includegraphics[width=\columnwidth]{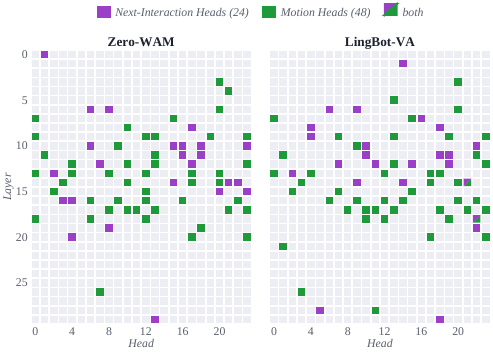}
  \caption{\textbf{Layer--head distribution of selected heads in \zerowam and LingBot-VA.}
    In both architectures, the identified Next-Interaction Heads and Motion Heads concentrate primarily in the middle transformer layers.}
  \label{fig:\currfilebase}
\end{figure}

\paragraph{Discovery procedure.}
To identify functional heads for a given WAM, we freeze the post-trained model and run a single teacher-forced forward pass over a small set of recorded demonstration episodes from its benchmark spanning diverse tasks, with only a few episodes per task.
Based on an attention analysis of the visual tokens, we compute both the next-interaction selectivity score $S_{\mathrm{N}}$ and the motion alignment score $S_{\mathrm{M}}$ for every video attention head.
We then rank all heads by each score and select the top $K_{\mathrm{N}}$ Next-Interaction Heads and the top $K_{\mathrm{M}}$ Motion Heads. The selected heads remain fixed and are reused across downstream tasks during adaptation.
Because discovery requires no backpropagation or parameter updates, it serves as a lightweight, one-time preprocessing step before fine-tuning.

\paragraph{Selected heads.}
\cref{fig:head_map} visualizes the spatial distribution of the $\Kni$ Next-Interaction Heads and $\Kmo$ Motion Heads across layers for both \zerowam and LingBot-VA.
In both architectures, the two functional families are primarily concentrated in the middle transformer layers.
They are completely disjoint in \zerowam and nearly disjoint in LingBot-VA.

\paragraph{Interaction targets and controls.}
Contact events (grasps, presses, releases) and their 3D locations are extracted from simulator states and projected onto the head-camera token grid.
The upcoming interaction target $\mathcal{T}$ for frame $f$ is the token of the earliest subsequent contact, plus the other arm's next contact token if it occurs within one frame; frames whose targets lie outside the view are skipped.
The \emph{previous} control is the most recent prior contact token, and the \emph{after-next} control is the first contact token after $\mathcal{T}$.
The \emph{random} control samples tokens outside the $3{\times}3$ neighborhood of $\mathcal{T}$, excluding gripper tokens and previous/after-next targets; the \emph{equidistant} control samples under the same exclusions from tokens whose gripper distance is within one token of that of $\mathcal{T}$.
Both match $\mathcal{T}$'s cardinality and are averaged over 10 draws; frames where the previous or after-next target is undefined or out of view are omitted from that control.
Log-ratios in \cref{eq:score_ni} are computed from frame-averaged $D_{\ell,h}$, then averaged across episodes with equal weight per task, before taking the minimum across controls.

\paragraph{Optical flow alignment.}
We estimate reference optical flow $\mathbf{u}^{f}$ using DIS~\cite{kroeger2016dis} between consecutive recorded head-camera frames ($f-1$ and $f$), and spatially average the flow within each token cell.
To focus on meaningful motion, we compute the expectation in \cref{eq:score_motion} over token cells with reference flow magnitudes exceeding $0.25$ tokens, average across frames within each episode, and then aggregate the results equally across tasks.
For visualization in \cref{fig:motion_flow,fig:supp_motion_tasks}, both the reference and attention-derived motion fields are anchored at token cells in frame $f$, with arrows pointing to their corresponding locations in frame $f-1$.

\begin{figure}[t]
  \centering
  \includegraphics[width=\columnwidth]{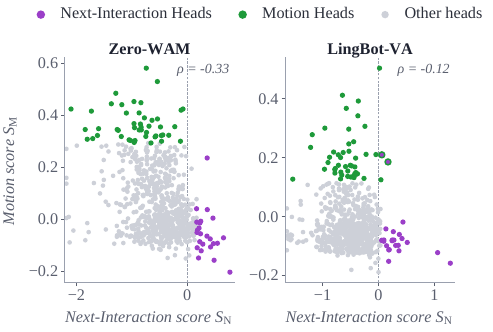}
  \caption{\textbf{Two distinct head families.}
    Next-Interaction score $S_{\mathrm{N}}$ versus Motion score $S_{\mathrm{M}}$ across all 720 video attention heads.
    In both architectures, the two families occupy opposite regions of the score space and exhibit a negative rank correlation $\rho$.
    Two LingBot-VA heads appearing in both sets are ringed.}
  \label{fig:\currfilebase}
\end{figure}

\paragraph{Separation of the two functional families.}
\cref{fig:supp_head_scatter} compares next-interaction selectivity $S_{\mathrm{N}}$ and motion alignment $S_{\mathrm{M}}$ across video attention heads.
In both models, Next-Interaction Heads exhibit high target selectivity but low motion alignment, while Motion Heads show the reverse.
This separation supports their distinct functional roles in interaction targeting and motion modeling.

\FloatBarrier
\subsection{Causal Head Ablations}
\label{sec:supp_ablation}

\cref{tab:supp_ablation_pertask} provides the per-task breakdown of the inference-time causal head ablations summarized in \cref{fig:head_ablation}.
In all ablations, we zero the outputs of the selected video heads on the robot's own video tokens while leaving the human demonstration conditioning and the action stream intact.
Across every unseen task, removing either Next-Interaction Heads or Motion Heads degrades closed-loop manipulation substantially more than removing equally many random heads.
\begin{table*}[t]
  \centering
  \small
  \setlength{\tabcolsep}{3pt}
  \renewcommand{\arraystretch}{1.1}
  \caption{\textbf{Per-task breakdown of causal head ablations} extending \cref{fig:head_ablation} on \zerowam after Full FT, reporting mean${\pm}$std over three seeds with 100 trials per seed.
    Across tasks, ablating $\mathcal{H}_{\mathrm{N}}^{v}$ or $\mathcal{H}_{\mathrm{M}}^{v}$ degrades success substantially more than random-head controls.}
  \label{tab:\currfilebase}
  \begin{tabular*}{\linewidth}{@{\extracolsep{\fill}}lrccccccc@{}}
    \toprule
    Removed heads & \#Heads
      & \mrh{Place object\\on scale} & \mrh{Stamp\\seal} & \mrh{Move stapler\\to pad}
      & \mrh{Place bread\\in basket} & \mrh{Place\\empty cup} & \mrh{Stack\\blocks three} & Average \\
    \midrule
    None & 0
      & \ms{43.33}{1.25} & \ms{73.33}{6.65} & \ms{85.33}{0.47}
      & \ms{42.67}{2.87} & \ms{91.33}{2.05} & \ms{11.33}{3.09} & \ms{57.89}{0.77} \\
    Random heads & 24
      & \ms{44.33}{0.47} & \ms{52.00}{3.74} & \ms{84.33}{3.86}
      & \ms{40.00}{4.32} & \ms{92.33}{1.25} & \ms{10.33}{2.05} & \ms{53.89}{1.65} \\
    Next-Interaction Heads & 24
      & \ms{26.00}{2.45} & \ms{38.00}{2.94} & \ms{64.67}{2.49}
      & \ms{24.33}{2.49} & \ms{90.33}{0.94} & \ms{0.33}{0.47} & \ms{40.61}{0.28} \\
    Random heads & 48
      & \ms{37.00}{1.41} & \ms{55.00}{2.45} & \ms{72.00}{2.16}
      & \ms{30.33}{3.40} & \ms{80.00}{3.27} & \ms{10.00}{2.94} & \ms{47.39}{1.29} \\
    Motion Heads & 48
      & \ms{0.67}{0.47} & \ms{0.67}{0.47} & \ms{3.33}{0.94}
      & \ms{0.67}{0.47} & \ms{0.33}{0.47} & \ms{0.00}{0.00} & \ms{0.94}{0.21} \\
    \bottomrule
  \end{tabular*}
\end{table*}

\FloatBarrier
\subsection{Causal Ablations on Other WAMs}
\label{sec:supp_crosswam}

We extend the causal ablations to four additional WAMs to examine whether the two head families play similar functional roles across architectures and benchmarks.

\paragraph{Models and benchmarks.}
We evaluate \fastwam~\cite{yuan2026fastwam} and \motus~\cite{bi2025motus} on \robotwin~2.0~\cite{chen2025robotwin}, \openwam~\cite{wang2026openwam} and \cosmosedge~\cite{nvidia2026cosmos3} on \libero~\cite{liu2023libero}.
The first three models use separate video and action experts, whereas \cosmosedge denoises video latents and actions in a shared stream.
All evaluations use released checkpoints without fine-tuning and cover tasks seen during training, testing the heads' causal roles in distribution.

\paragraph{Protocol.}
We identify both head families using the procedure in \cref{sec:supp_discovery} with a few demonstrations per task.
Head budgets are guided by offline loss comparisons against layer-matched random ablations, following \cref{fig:head_budget}.
The resulting $K_{\mathrm{N}}/K_{\mathrm{M}}$ are $30/48$ for \fastwam, $96/48$ for \motus, $48/96$ for \openwam, and $30/19$ for \cosmosedge.
At inference time, we zero the selected heads' outputs on video tokens and compare against equally many uniformly sampled random heads on identical episodes (5,000 each for \fastwam and \motus, and 2,000 each for \openwam and \cosmosedge), following each model's official evaluation protocol.
The unablated success rates in \cref{tab:supp_crosswam} are published results.

Two architectures require minor adjustments.
Since \fastwam encodes only the current frame at inference time, we identify its Motion Heads using training-time video prediction and ablate them during current-frame encoding.
For \cosmosedge, we measure motion alignment from the current to the last predicted latent frame and restrict ablation to video-latent tokens in the shared stream.
\begin{table*}[t]
  \centering
  \small
  \setlength{\tabcolsep}{4pt}
  \renewcommand{\arraystretch}{1.1}
  \caption{\textbf{Causal head ablation on four additional public WAMs.}
    On each released checkpoint, we rediscover the two head families (\cref{sec:supp_discovery}) and zero them on the
    video tokens at inference time, without fine-tuning, against the same number of random video heads on the same
    episodes.
    Original SR: published success rate (\%) of the unablated checkpoint over all tasks of the benchmark; the ablation
    columns give the success rate (\%) after zeroing the random or the selected heads.}
  \label{tab:\currfilebase}
  \begin{tabular*}{\linewidth}{@{\extracolsep{\fill}}llccc@{}}
    \toprule
    \multicolumn{5}{@{}l}{\textbf{(a) Ablating Next-Interaction Heads}} \\
    \midrule
    Model & Benchmark & Original SR & \mrh{Random head\\ablation} & \mrh{Next-Interaction\\head ablation} \\
    \midrule
    \fastwam~\cite{yuan2026fastwam}         & \robotwin~2.0 & 91.9 & 90.0       & 85.6 \\
    \motus~\cite{bi2025motus}               & \robotwin~2.0 & 88.7 & 81.7       & 74.9 \\
    \openwam~\cite{wang2026openwam}         & \libero~\cite{liu2023libero} & 99.3 & 98.6       & 91.9 \\
    \cosmosedge~\cite{nvidia2026cosmos3}    & \libero~\cite{liu2023libero} & 98.0 & 81.9       & 51.7 \\
    \midrule
    \multicolumn{5}{@{}l}{\textbf{(b) Ablating Motion Heads}} \\
    \midrule
    Model & Benchmark & Original SR & \mrh{Random head\\ablation} & \mrh{Motion head\\ablation} \\
    \midrule
    \fastwam~\cite{yuan2026fastwam}         & \robotwin~2.0 & 91.9 & 91.8       & 90.5 \\
    \motus~\cite{bi2025motus}               & \robotwin~2.0 & 88.7 & 85.0       & 75.8 \\
    \openwam~\cite{wang2026openwam}         & \libero~\cite{liu2023libero} & 99.3 & 98.2       & 72.4 \\
    \cosmosedge~\cite{nvidia2026cosmos3}    & \libero~\cite{liu2023libero} & 98.0 & 96.6       & 37.5 \\
    \bottomrule
  \end{tabular*}
\end{table*}

\paragraph{Results.}
As shown in \cref{tab:supp_crosswam}, ablating the selected heads leads to larger success drops than ablating equally many random heads across the four additional WAMs. The magnitude of these effects varies depending on how each model uses video information to predict action.

\emph{Next-Interaction Heads} play an important role in action prediction across architectures.
Ablating them produces clear drops in all four models.
\emph{Motion Heads} matter most when actions depend on predicted future frames. Their impact is therefore limited in \fastwam, which does not generate future frames at inference: Actions are denoised from current-frame features computed once by the video expert, leaving no consecutive frames for \emph{Motion Heads} to relate.
In contrast, \openwam, \motus, and \cosmosedge jointly denoise future frames and actions, so errors in the predicted motion carry over to the actions and yield larger ablation drops.
In \zerowam, which generates future video before decoding actions, Motion Head ablation reduces success to nearly zero (\cref{fig:head_ablation}).

\subsection{Consistency and Generality of Selected Heads}
\label{sec:supp_consistency}

\begin{figure*}[p]
  \centering
  \begin{subfigure}[b]{0.5\textwidth}
    \centering
    \includegraphics[width=\linewidth]{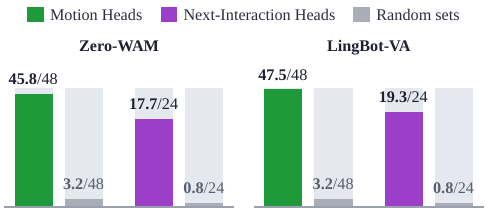}
    \caption{Consistency of head selection.}
    \label{fig:supp_head_overlap}
  \end{subfigure}\hfill
  \begin{subfigure}[b]{0.325\textwidth}
    \centering
    \includegraphics[width=\linewidth]{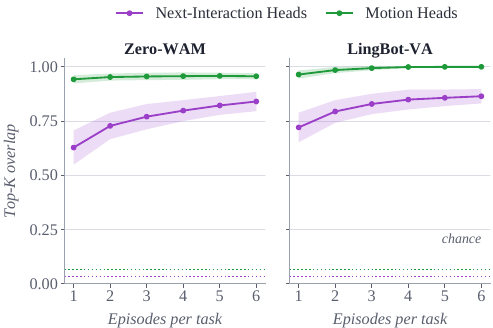}
    \caption{Discovery data size.}
    \label{fig:supp_head_datasize}
  \end{subfigure}
  \caption{\textbf{Stability of head selection.}
    (a)~Mean number of heads shared by the top-$K$ sets selected from two disjoint subsets of the discovery episodes, over 200 random partitions into five subsets.
    Gray bars show the number shared by two random sets of $K$ heads, and light bars mark $K$.
    (b)~Overlap between top-$K$ head sets selected from two disjoint demonstration sets with the same number of episodes per task, reported as mean${\pm}$std over 200 draws with dotted lines showing random-set chance levels.
    Motion Heads stabilize from a single episode per task, while Next-Interaction Heads converge steadily as more episodes are added.}
  \label{fig:\currfilebase}
\end{figure*}

We examine the stability of the two head families in \zerowam and LingBot-VA, following the discovery procedure in \cref{sec:discovery}.

\paragraph{Stability across demonstration subsets and data sizes.}
Following~\cite{mitra2026robotsteering}, we repeat head selection on disjoint demonstration subsets and measure the overlap between the selected sets.
As shown in \cref{fig:supp_head_overlap}, both families exhibit substantially greater overlap than random selections.
Motion Heads remain stable even with a single demonstration per task, while Next-Interaction Heads become increasingly consistent as more demonstrations are used (\cref{fig:supp_head_datasize}).

\subsection{Visualizing Functional Specialization}
\label{sec:supp_interp}

\providecommand{\hlsuppfont}{\fontsize{6.4}{7}\selectfont\bfseries}%
\providecommand{\hlsuppgroup}[4]{%
  \put(#1,#2){\makebox(0,#3){\textcolor{hlline}{\rule{0.72\unitlength}{#3\unitlength}}}}%
  \put(#1,#2){\makebox(0,#3){\rotatebox{90}{\setlength{\fboxsep}{0pt}\colorbox{white}{\hspace{5\unitlength}%
    \raisebox{0pt}[4.3pt][0pt]{\hlsuppfont\textcolor{hlink}{#4}}\hspace{5\unitlength}}}}}}
\definecolor{hlline}{HTML}{B8B8B8}
\definecolor{hlink}{HTML}{1E2230}
\begin{figure*}[p]
  \centering
  \setlength{\unitlength}{\dimexpr\textwidth*95/100/495\relax}%
  \begin{picture}(495,187.89)
    \hlsuppgroup{5.3}{68.29}{119.6}{Zero-WAM}
    \put(13,129.34){\makebox(16,58.55){\rotatebox{90}{\shortstack{\hlsuppfont\textcolor{hlink}{Move stapler}\\\hlsuppfont\textcolor{hlink}{to pad}}}}}
    \put(31,129.34){\includegraphics[viewport=21.6 71.973 495 131.707,clip,width=464\unitlength]{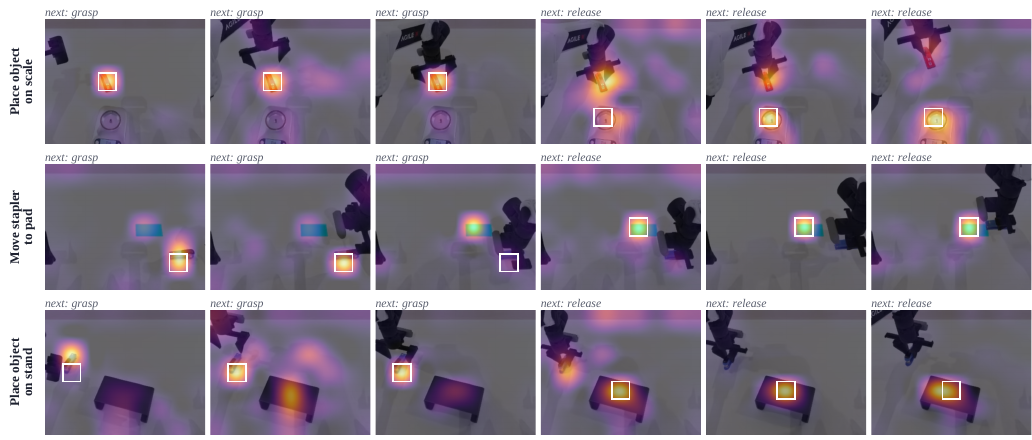}}
    \put(13,68.29){\makebox(16,58.55){\rotatebox{90}{\shortstack{\hlsuppfont\textcolor{hlink}{Place object}\\\hlsuppfont\textcolor{hlink}{on stand}}}}}
    \put(31,68.29){\includegraphics[viewport=21.6 2.16 495 61.893,clip,width=464\unitlength]{fig/heads/supp_ni_time.pdf}}
    \hlsuppgroup{5.3}{0}{60.29}{LingBot-VA}
    \put(13,0){\makebox(16,60.29){\rotatebox{90}{\shortstack{\hlsuppfont\textcolor{hlink}{Stack blocks}\\\hlsuppfont\textcolor{hlink}{three}}}}}
    \put(31,0){\includegraphics[viewport=0 0.72 495 65.04,clip,width=464\unitlength]{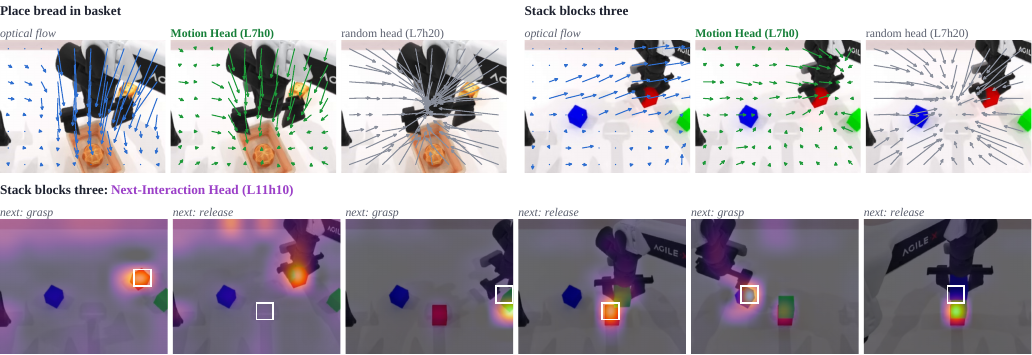}}
  \end{picture}
  \caption{\textbf{Next-Interaction Heads over an episode.}
    Attention maps of a Next-Interaction Head in \zerowam (L10h23, six consecutive frames spaced 0.32\,s apart) and in LingBot-VA (L11h10, every fifth frame of the episode), with each map scaled to its own maximum as in \cref{fig:ni_attention}.
    The white square marks the upcoming interaction target $\mathcal{T}$.
    In \zerowam, attention focuses on the object to be grasped and, once it is grasped, shifts toward the upcoming release location; in LingBot-VA, it likewise concentrates on the upcoming target at key transitions.}
  \label{fig:\currfilebase}
\end{figure*}

\providecommand{\hlsuppfont}{\fontsize{6.4}{7}\selectfont\bfseries}%
\providecommand{\hlsuppgroup}[4]{%
  \put(#1,#2){\makebox(0,#3){\textcolor{hlline}{\rule{0.72\unitlength}{#3\unitlength}}}}%
  \put(#1,#2){\makebox(0,#3){\rotatebox{90}{\setlength{\fboxsep}{0pt}\colorbox{white}{\hspace{5\unitlength}%
    \raisebox{0pt}[4.3pt][0pt]{\hlsuppfont\textcolor{hlink}{#4}}\hspace{5\unitlength}}}}}}
\definecolor{hlline}{HTML}{B8B8B8}
\definecolor{hlink}{HTML}{1E2230}
\begin{figure*}[p]
  \centering
  \setlength{\unitlength}{\dimexpr\textwidth*95/100/495\relax}%
  \begin{picture}(495,162.92)
    \hlsuppgroup{5.3}{86.68}{75.7}{Zero-WAM}
    \put(13.5,86.68){\includegraphics[viewport=0 82.62 243.18 161,clip,width=236.55\unitlength]{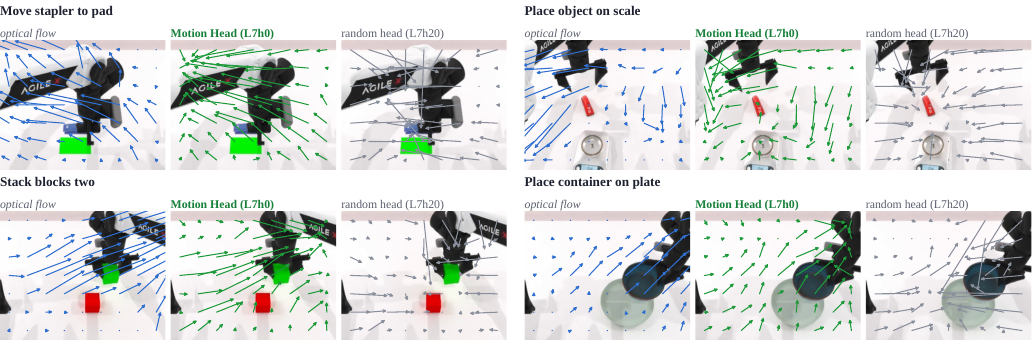}}
    \put(258.45,86.68){\includegraphics[viewport=0 0.72 243.18 79.1,clip,width=236.55\unitlength]{fig/heads/supp_motion_tasks.pdf}}
    \hlsuppgroup{5.3}{0}{77.4}{LingBot-VA}
    \put(13.5,0){\includegraphics[viewport=0 87.264 243.18 170.208,clip,width=236.55\unitlength]{fig/heads/supp_lingbot.pdf}}
    \put(258.45,0){\includegraphics[viewport=251.82 87.264 495 170.208,clip,width=236.55\unitlength]{fig/heads/supp_lingbot.pdf}}
  \end{picture}
  \caption{\textbf{Motion Heads on additional tasks.}
    Reference optical flow $\mathbf{u}^{f}$ and attention-induced motion fields $\widehat{\mathbf{u}}_{\ell,h}^{f}$ of the Motion Head L7h0 and the random head L7h20 from \cref{fig:motion_flow} on two tasks per model; L7h0 is a selected Motion Head in both models, L7h20 in neither.
    The Motion Head closely tracks the moving arm and manipulated objects in every example, whereas the random head does not.}
  \label{fig:\currfilebase}
\end{figure*}

Across additional tasks, the same Next-Interaction Head shifts its attention from grasp targets to subsequent release targets as interactions unfold (\cref{fig:supp_ni_time}), while the same Motion Head produces attention-induced motion fields that align with reference optical flow (\cref{fig:supp_motion_tasks}).
Both figures show the same patterns in LingBot-VA, providing qualitative evidence of both specializations in a unified, language-conditioned WAM.

\section{Additional Experimental Results}
\label{sec:supp_results}

\begin{table*}[t]
  \centering
  \small
  \setlength{\tabcolsep}{3pt}
  \renewcommand{\arraystretch}{1.1}
  \caption{\textbf{Per-task breakdown of pathway complementarity} corresponding to \cref{tab:complementarity} on the six unseen \robotwin tasks with \zerowam, reporting mean${\pm}$std over three seeds with 100 trials per seed.
    Combining both pathways in \method achieves the best or tied-best performance on five of six tasks. Best per column in bold.}
  \label{tab:\currfilebase}
  \begin{tabular*}{\linewidth}{@{\extracolsep{\fill}}lccccccc@{}}
    \toprule
    Adapted heads
      & \mrh{Place object\\on scale} & \mrh{Stamp\\seal} & \mrh{Move stapler\\to pad}
      & \mrh{Place bread\\in basket} & \mrh{Place\\empty cup} & \mrh{Stack\\blocks three} & Average \\
    \midrule
    None (\zerowam~\cite{zhou2026zerowam})
      & \ms{24.67}{2.05} & \ms{47.00}{4.55} & \ms{69.14}{2.93}
      & \ms{35.00}{3.74} & \ms{84.87}{0.18} & \ms{9.00}{2.16} & \ms{44.95}{0.79} \\
    All (Full FT)
      & \ms{43.33}{1.25} & \msb{73.33}{6.65} & \ms{85.33}{0.47}
      & \ms{42.67}{2.87} & \ms{91.33}{2.05} & \msb{11.33}{3.09} & \ms{57.89}{0.77} \\
    \midrule
    Next-Interaction Heads only
      & \ms{44.33}{1.70} & \ms{59.00}{4.55} & \ms{86.00}{0.82}
      & \ms{48.33}{2.87} & \ms{91.67}{3.09} & \ms{7.67}{0.47} & \ms{56.17}{0.14} \\
    \quad random heads control
      & \ms{38.33}{7.72} & \ms{61.00}{2.94} & \ms{80.00}{2.94}
      & \ms{40.00}{5.35} & \ms{86.67}{2.62} & \ms{7.33}{0.47} & \ms{52.22}{0.44} \\
    Motion Heads only
      & \ms{41.67}{5.56} & \ms{62.67}{6.65} & \ms{86.00}{2.16}
      & \ms{45.33}{4.50} & \ms{92.67}{0.47} & \ms{8.33}{3.09} & \ms{56.11}{1.36} \\
    \quad random heads control
      & \ms{40.00}{1.63} & \ms{59.33}{8.73} & \ms{84.00}{0.82}
      & \ms{44.67}{2.87} & \ms{91.33}{1.70} & \ms{6.67}{0.47} & \ms{54.33}{1.65} \\
    \midrule
    Both (random heads control)
      & \ms{45.33}{3.30} & \ms{68.67}{5.56} & \ms{84.33}{2.49}
      & \ms{45.67}{3.40} & \ms{92.67}{2.05} & \ms{7.33}{1.25} & \ms{57.33}{1.89} \\
    Both (\method)
      & \msb{54.00}{5.72} & \msb{73.33}{3.40} & \msb{89.00}{1.41}
      & \msb{52.67}{4.03} & \msb{97.00}{1.63} & \ms{7.67}{1.70} & \msb{62.28}{2.30} \\
    \bottomrule
  \end{tabular*}
\end{table*}

\begin{figure*}[t]
  \centering
  \includegraphics[width=\textwidth]{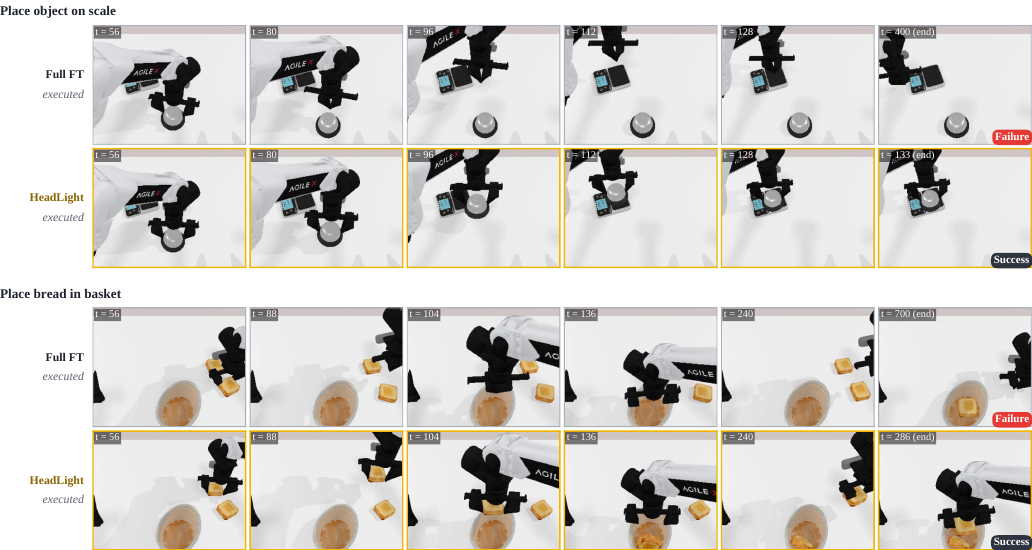}
  \caption{\textbf{Same-scene rollouts on unseen \robotwin tasks with \zerowam.}
  Executed head-camera frames at matching action steps $t$, with the final column showing the end of each episode.
  Full FT closes its gripper on or beside the object but fails to lift it.
  \method successfully grasps, carries, and places the object in both scenes.}
  \label{fig:\currfilebase}
\end{figure*}

\begin{table*}[tp]
  \centering
  \small
  \setlength{\tabcolsep}{3pt}
  \renewcommand{\arraystretch}{1.1}
  \caption{\textbf{Online head-budget ablation per task} on unseen \robotwin tasks with \zerowam.
    The first row is the unadapted model.
    Each other row adapts only the top-$K$ Next-Interaction Heads in the video stream or only the action heads paired with the top-$K$ Motion Heads, reporting mean${\pm}$std over three seeds with 100 trials per seed.
    Success peaks at $K_{\mathrm{N}}{=}\Kni$ and rises up to $K_{\mathrm{M}}{=}\Kmo$.}
  \label{tab:\currfilebase}
  \begin{tabular*}{\linewidth}{@{\extracolsep{\fill}}lrrccccccc@{}}
    \toprule
    Adapted heads & $K$ & \#Params
      & \mrh{Place object\\on scale} & \mrh{Stamp\\seal} & \mrh{Move stapler\\to pad}
      & \mrh{Place bread\\in basket} & \mrh{Place\\empty cup} & \mrh{Stack\\blocks three} & Average \\
    \midrule
    None (\zerowam~\cite{zhou2026zerowam}) & 0 & --
      & \ms{24.67}{2.05} & \ms{47.00}{4.55} & \ms{69.14}{2.93} & \ms{35.00}{3.74} & \ms{84.87}{0.18} & \ms{9.00}{2.16} & \ms{44.95}{0.79} \\
    \midrule
    Next-Interaction Heads & 12 & 18.9M
      & \ms{39.00}{2.16} & \ms{53.33}{5.31} & \ms{77.67}{3.77} & \ms{38.33}{2.36} & \ms{82.33}{0.94} & \ms{9.67}{2.05} & \ms{50.06}{1.85} \\
                           & 24 & 37.8M
      & \ms{44.33}{1.70} & \ms{59.00}{4.55} & \ms{86.00}{0.82} & \ms{48.33}{2.87} & \ms{91.67}{3.09} & \ms{7.67}{0.47} & \ms{56.17}{0.14} \\
                           & 36 & 56.6M
      & \ms{43.67}{4.71} & \ms{52.67}{3.68} & \ms{84.33}{2.49} & \ms{38.67}{4.92} & \ms{92.00}{1.63} & \ms{5.67}{1.25} & \ms{52.83}{0.49} \\
                           & 48 & 75.5M
      & \ms{42.67}{6.94} & \ms{56.33}{5.44} & \ms{84.67}{1.89} & \ms{44.00}{2.16} & \ms{94.67}{2.87} & \ms{8.00}{0.82} & \ms{55.06}{1.98} \\
    \midrule
    Motion Heads           & 12 & 18.9M
      & \ms{34.00}{2.16} & \ms{57.67}{2.49} & \ms{80.33}{0.47} & \ms{42.67}{3.77} & \ms{91.00}{3.56} & \ms{3.00}{0.82} & \ms{51.44}{1.09} \\
                           & 24 & 37.8M
      & \ms{38.67}{1.70} & \ms{60.67}{5.73} & \ms{80.33}{1.70} & \ms{47.00}{5.72} & \ms{93.33}{1.25} & \ms{4.67}{1.25} & \ms{54.11}{1.10} \\
                           & 36 & 56.6M
      & \ms{37.67}{2.62} & \ms{64.00}{4.32} & \ms{82.33}{0.47} & \ms{46.33}{2.49} & \ms{92.33}{0.94} & \ms{6.67}{0.47} & \ms{54.89}{0.67} \\
                           & 48 & 75.5M
      & \ms{41.67}{5.56} & \ms{62.67}{6.65} & \ms{86.00}{2.16} & \ms{45.33}{4.50} & \ms{92.67}{0.47} & \ms{8.33}{3.09} & \ms{56.11}{1.36} \\
    \bottomrule
  \end{tabular*}
\end{table*}

\subsection{Per-Task Pathway Complementarity}
\label{sec:supp_pertask}

\cref{tab:supp_complementarity_pertask} breaks down \cref{tab:complementarity} per task.
Neither pathway alone matches the joint combination: \method achieves the highest or tied-highest success on five of six unseen tasks.

\subsection{Head Budgets}
\label{sec:supp_budget}

\cref{tab:supp_head_budget_pertask} reports unseen-task success when varying $K$ within each pathway, complementing \cref{fig:head_budget}.
Next-Interaction Head success peaks at $K_{\mathrm{N}}{=}24$ and does not improve at 36 or 48, consistent with larger budgets increasingly perturbing video prediction.
Motion Head success improves steadily up to $K_{\mathrm{M}}{=}48$, supporting the default budgets.

\subsection{Qualitative Comparison}
\label{sec:supp_qualitative}

\cref{fig:supp_method_compare} compares closed-loop rollouts of Full FT and \method on identical unseen scenes.
In both examples, the Full FT gripper closes without lifting the object and reaches the target empty-handed, despite its predicted video completing the grasp and transport.
\method successfully grasps, transports, and places the objects in both scenes.

\section{Real-Robot Experiments}
\label{sec:supp_real}

\subsection{Hardware and Data Collection}
\label{sec:supp_real_hw}

\begin{table*}[tp]
  \centering
  \small
  \setlength{\tabcolsep}{4pt}
  \renewcommand{\arraystretch}{1.1}
  \caption{\textbf{Real-robot tasks.}
    Each unseen task recombines the objects and manipulation steps of a seen task into a novel goal.
    A rollout succeeds when the final scene satisfies the specified success condition.}
  \label{tab:supp_real_tasks}
  \begin{tabular*}{\linewidth}{@{\extracolsep{\fill}}lll@{}}
    \toprule
    Split & Task & Success condition \\
    \midrule
    Seen   & Strawberry plate  & strawberry plush on the blue plate \\
           & Red cube on plate & red cube on the blue plate \\
           & Two plushies      & lemon plush out of the bowl, strawberry plush in the bowl \\
           & Cube drawer       & red cube out of the drawer on the table, drawer closed \\
    \midrule
    Unseen & Swap plushies     & strawberry plush out of the bowl, lemon plush in the bowl \\
           & Lemon plate       & lemon plush on the blue plate \\
           & Lemon drawer      & lemon plush out of the drawer on the table, drawer closed \\
           & Strawberry drawer & strawberry plush out of the drawer on the table, drawer closed \\
    \bottomrule
  \end{tabular*}
\end{table*}

The real-robot experiments in \cref{sec:real_world} are carried out on a 7-DoF Franka Panda arm equipped with a parallel-jaw gripper, observed by a third-person ZED~2i camera and a wrist-mounted ZED~Mini camera as shown in \cref{fig:real_setup}.
All eight tasks share a single tabletop workspace and a common set of objects: a red cube, a strawberry plush, a lemon plush, a blue plate, a bowl, and a drawer box.
We collect 100 teleoperated demonstrations for each of the four seen tasks, and every method is adapted on the same 400 demonstrations.
\method reuses the heads discovered on \robotwin for each WAM.
Consistent with the simulation setup in \cref{sec:exp_setup}, LingBot-VA is conditioned on task language instruction and \zerowam is conditioned on a human demonstration video of the task.

At control time, policies output absolute end-effector poses relative to the initial episode pose along with a binary gripper command at 13.5\,Hz.
\zerowam processes both camera views at $192{\times}352$ resolution and predicts action chunks spanning 16 actions and 8 video frames, corresponding to 1.2\,s per chunk.
Human demonstration videos for \zerowam are recorded from the third-person camera and resampled to match the frame rate of the robot observations.

\subsection{Tasks and Evaluation Protocol}
\label{sec:supp_real_tasks}

\begin{table*}[tp]
  \centering
  \setlength{\tabcolsep}{2.5pt}
  \renewcommand{\arraystretch}{1.1}
  \caption{\textbf{Real-robot manipulation on a Franka Panda arm.}
    Policies adapted on four seen tasks with 100 demonstrations each are evaluated over 20 rollouts per task on both seen tasks and four unseen object--goal recombinations, matching the averages in \cref{fig:real_setup}.
    \method consistently outperforms Full FT and LoRA across both WAMs, with the largest gains on unseen tasks.
    Best per backbone in bold.}
  \label{tab:\currfilebase}
  \begin{tabular*}{\linewidth}{@{\extracolsep{\fill}}lcccccccccc@{}}
    \toprule
    & \multicolumn{5}{c}{Seen tasks} & \multicolumn{5}{c}{Unseen tasks} \\
    \cmidrule(lr){2-6}\cmidrule(l){7-11}
    Method
      & \mrh{Strawberry\\plate} & \mrh{Red cube\\on plate} & \mrh{Two\\plushies} & \mrh{Cube\\drawer} & \mrh{Avg.\\(\%)}
      & \mrh{Swap\\plushies} & \mrh{Lemon\\plate} & \mrh{Lemon\\drawer} & \mrh{Strawberry\\drawer} & \mrh{Avg.\\(\%)} \\
    \midrule
    \zerowam~\cite{zhou2026zerowam} + Full FT
      & 18/20 & 17/20 & \textbf{16/20} & 13/20 & 80.00
      & 7/20 & 13/20 & 10/20 & 9/20 & 48.75 \\
    \zerowam~\cite{zhou2026zerowam} + LoRA
      & 17/20 & 17/20 & 15/20 & 13/20 & 77.50
      & 8/20 & 15/20 & \textbf{13/20} & 12/20 & 60.00 \\
    \zerowam~\cite{zhou2026zerowam} + \method
      & \textbf{19/20} & \textbf{20/20} & 15/20 & \textbf{16/20} & \textbf{87.50}
      & \textbf{10/20} & \textbf{18/20} & \textbf{13/20} & \textbf{14/20} & \textbf{68.75} \\
    \midrule
    LingBot-VA~\cite{li2026causalwam} + Full FT
      & 18/20 & 16/20 & 11/20 & 14/20 & 73.75
      & 10/20 & 11/20 & 9/20 & 10/20 & 50.00 \\
    LingBot-VA~\cite{li2026causalwam} + LoRA
      & 17/20 & 15/20 & 10/20 & \textbf{15/20} & 71.25
      & 11/20 & 14/20 & 11/20 & \textbf{13/20} & 61.25 \\
    LingBot-VA~\cite{li2026causalwam} + \method
      & \textbf{20/20} & \textbf{19/20} & \textbf{13/20} & \textbf{15/20} & \textbf{83.75}
      & \textbf{14/20} & \textbf{17/20} & \textbf{14/20} & \textbf{13/20} & \textbf{72.50} \\
    \bottomrule
  \end{tabular*}
\end{table*}

\begin{figure*}[p]
  \centering
  \includegraphics[width=\linewidth]{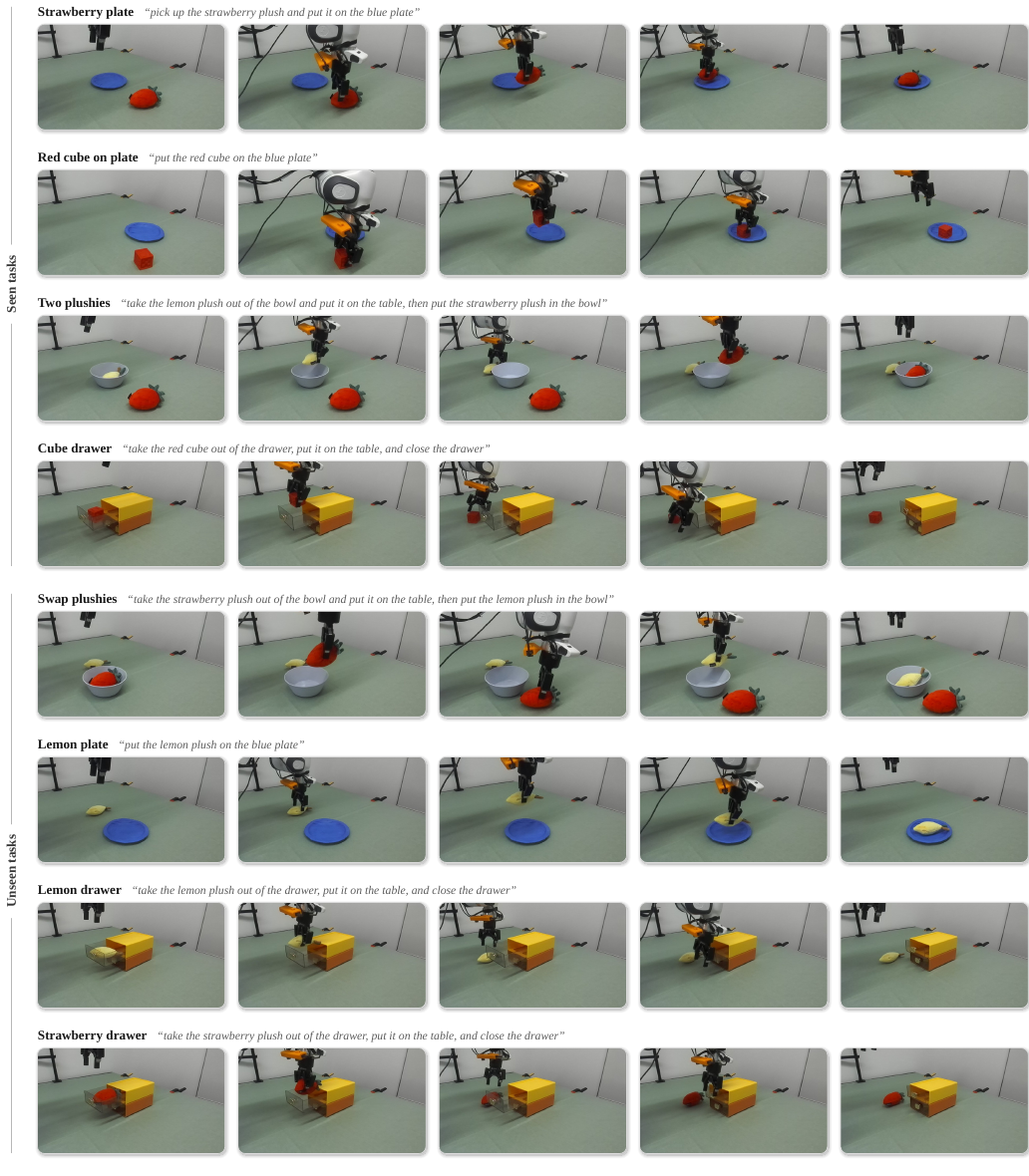}
  \caption{\textbf{Real-robot task rollouts.}
    Third-person keyframes ordered over time from left to right together with language instructions for all eight tasks in \cref{tab:real_robot}.
    Seen tasks provide teleoperated adaptation demonstrations, whereas unseen tasks recombine the same tabletop objects into novel manipulation goals.}
  \label{fig:\currfilebase}
\end{figure*}

\begin{figure*}[b]
  \centering
  \includegraphics[width=\textwidth]{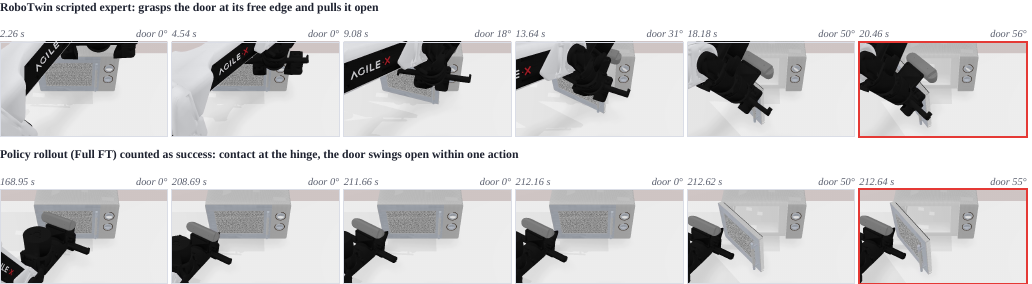}
  \caption{\textbf{Open Microwave counts successes without the intended interaction.}
    \emph{Top:} the \robotwin scripted expert grasps the free edge of the door and pulls it open.
    \emph{Bottom:} a Full FT rollout on the same scene touches the hinge-side corner of the door; the lightweight, undamped door swings from $0^\circ$ to $50^\circ$ within a single action and triggers the success check, highlighted in red.}
  \label{fig:\currfilebase}
\end{figure*}

\cref{tab:supp_real_tasks} defines the eight real-robot tasks, and \cref{fig:real_tasks} displays keyframes of a successful rollout for each task together with its language instruction.
The four seen tasks provide the teleoperated demonstrations used for adaptation, whereas the four unseen tasks have no robot demonstrations and instead recombine the same objects and manipulation primitives into novel goals.
A rollout is counted as successful if, once the robot stops, every object named in the instruction rests at its designated target location on the plate, in the bowl, or on the table, and for drawer tasks the drawer is pushed closed.
Each method is evaluated over 20 physical rollouts per task.

\subsection{Results}
\label{sec:supp_real_results}

\cref{tab:real_robot} reports the full success counts per-task of the real-robot corresponding to the averages shown in \cref{fig:real_setup}.
\method achieves the highest average success rate on both seen and unseen splits with each of \zerowam and LingBot-VA.
Its advantages are most pronounced on unseen object--goal recombinations, where \method outperforms LoRA by 8.75 percentage points on \zerowam and by 11.25 percentage points on LingBot-VA, confirming that functionally guided head adaptation transfers reliably to physical manipulation.

\newpage
\section{Notes on the Simulation Benchmark}
\label{sec:supp_sim}

\paragraph{Exclusion of Open Microwave.}
\zerowam~\cite{zhou2026zerowam} originally evaluated seven unseen \robotwin tasks, including \emph{Open Microwave}.
We exclude it because the simulator judges success solely by the hinge angle of the door: the simulated door is lightweight and undamped, so incidental contact near the hinge or the top edge can swing it open without grasping the handle (\cref{fig:supp_microwave}).
Replaying logged rollouts confirms that roughly 90\% of successes across adapted policies exploit this artifact.
The unadapted \zerowam average in \cref{tab:main_results} is recomputed on the remaining six tasks from the published per-task scores (44.95\% vs.\ 46.95\% across all seven).

\FloatBarrier

\end{document}